\documentclass{article}

    \PassOptionsToPackage{numbers}{natbib}

\usepackage[preprint]{neurips_2026}

\usepackage[utf8]{inputenc} %
\usepackage[T1]{fontenc}    %
\usepackage{hyperref}       %
\usepackage{url}            %
\usepackage{booktabs}       %
\usepackage{amsfonts, amsmath}       %
\usepackage{nicefrac}       %
\usepackage{wrapfig}
\usepackage{microtype}      %
\usepackage[dvipsnames]{xcolor}
\usepackage{dsfont}
\usepackage{subcaption}
\usepackage{color, colortbl}
\usepackage{float}
\usepackage{tcolorbox} %
\usepackage{graphicx,array}
\usepackage{enumitem} 
\usepackage{rotating}
\usepackage{multirow}
\usepackage{xspace}
\usepackage{soul}
\newcommand{\cv}[0]{\texttt{CV}}
\newcommand{\acc}[0]{\texttt{acc}}

\newcommand{\mae}[0]{\texttt{MAE}}
\newcommand{\mic}[0]{\texttt{MIC}}
\newcommand{\micx}[0]{\texttt{MIC}\xspace}
\newcommand{\mf}[0]{\texttt{MF}}

\newcommand{\dtest}[0]{D_{\text{test}}}
\newcommand{\dtrain}[0]{D_{\text{train}}}
\newcommand{\conf}[0]{\texttt{conf}_M}
\newcommand{\cmfg}[0]{\texttt{cMFG}}
\newcommand{\cmfgx}[0]{\texttt{cMFG}\xspace}
\newcommand{\imae}[0]{\texttt{ID-MAE}}
\newcommand{\imaex}[0]{\texttt{ID-MAE}\xspace}
\newcommand{\cmae}[0]{\texttt{CD-MAE}}
\newcommand{\cmaex}[0]{\texttt{CD-MAE}\xspace}

\newcommand{\dcv}[0]{\texttt{D-AvgCV}}
\newcommand{\mcv}[0]{\texttt{M-AvgCV}}
\newcommand{\dcvx}[0]{\texttt{D-AvgCV}\xspace}
\newcommand{\mcvx}[0]{\texttt{M-AvgCV}\xspace}
\newcommand{\mac}[0]{\texttt{MAC}}
\newcommand{\macx}[0]{\texttt{MAC}\xspace}
\newcommand{\mcc}[0]{\texttt{MCC}}
\newcommand{\mccx}[0]{\texttt{MCC}\xspace}
\newcommand{\mrc}[0]{\texttt{MRC}}
\newcommand{\mrcx}[0]{\texttt{MRC}\xspace}

\newcommand{\blue}[1]{\textcolor{blue}{#1}}
\newcommand{\sbrace}[1]{\left[#1\right]}

\title{Can LLMs Use Linguistic Uncertainty Markers to Reliably Reflect Intrinsic Confidence?}

\author{%
  \textbf{Gabrielle Kaili-May Liu}\qquad
  \textbf{Arman Cohan}\\\\
  Yale University\\
  {\small \texttt{\{kaili.liu, arman.cohan\}@yale.edu}} \\
}

\begin{document}

\maketitle

\begin{abstract}
LLMs’ linguistically expressed confidence should faithfully reflect their intrinsic uncertainty.
While recent work shows LLMs struggle to use epistemic markers (e.g., ``it is likely...'') in a human-aligned fashion,
it remains unclear whether models can apply their own linguistic confidence framework to associate markers with specific confidence levels in a stable and generalizable way, and how contextual features 
impact this ability. We conduct the first systematic study of this question, formalizing \textit{marker internal confidence} (\mic) as the estimated intrinsic confidence a model associates with a specific epistemic marker in a given task domain. We present 7 metrics to evaluate the stability of \mic s within and across distributions.
Applying our analysis framework to diverse models and tasks, we find that 
LLMs 
remain faithfully miscalibrated even under model-centric interpretation of marker meanings, struggling to differentiate markers by internal confidence across distributions despite preserving a somewhat consistent ranking order
across tasks. 
This supplies critical, complementary evidence to existing work toward a holistic understanding of faithful calibration in LLMs, emphasizing the need for more aligned and stable marker use to improve trustworthiness and reliability.\footnote{We release our code at \url{https://github.com/yale-nlp/marker_internal_confidence}.}
\end{abstract}

\section{Introduction} \label{sec:intro}
Understanding the reliability and trustworthiness of LLMs has become increasingly important as models are deployed across applications such as scientific discovery \cite{song2025evaluatinglargelanguagemodels, zhang2025advancingscientificmethodlarge}, medical diagnosis \cite{Johnson2023AssessingTA, zhou2025large}, and legal consulting \cite{dahl2024largelegalfictions, li-etal-2025-legalagentbench}. Yet despite their unprecedented capabilities, LLMs continue to suffer from hallucinations \cite{tonmoy2024comprehensive, 10.1145/3703155}, wherein false claims are habitually conveyed using decisive, assertive language \cite{xiao-wang-2021-hallucination, zhou-etal-2023-navigating, simhi2025trust}. In downstream settings, the risks of such misalignment are well-documented \cite{10.1145/3630106.3658941, zhou-etal-2024-relying}, with over- or under-confident generations able to mislead users, undermine trust, and foster over-reliance in LLM-based systems. Measuring the uncertainty of models has therefore become a primary means to guide AI-assisted decision-making and reduce potential harm \cite{zhou-etal-2024-relying, Steyvers_2025}.

\begin{figure}[t]
    \centering
    \includegraphics[width=0.65\columnwidth]{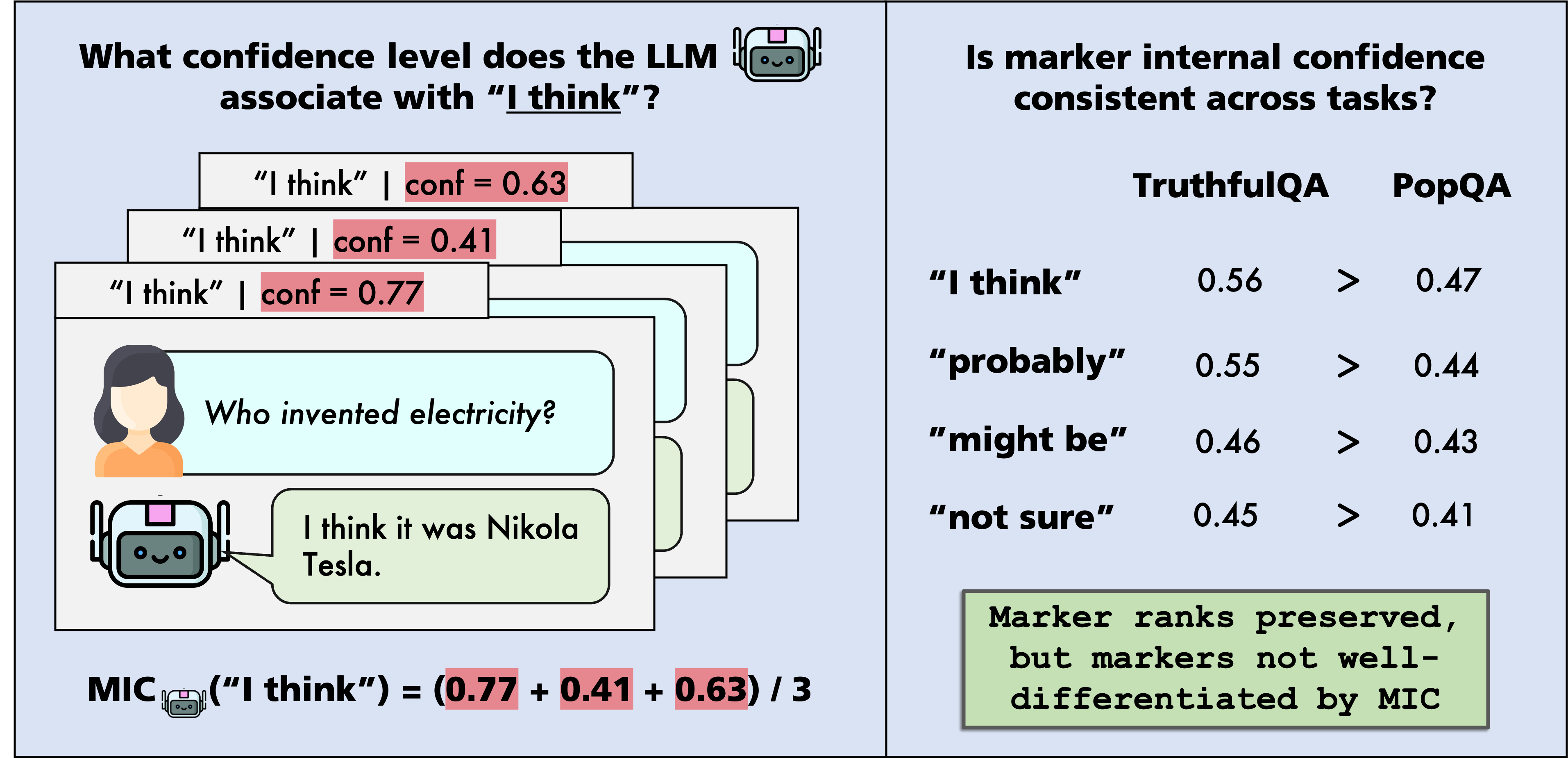}
    \caption{An example of our framework to calculate the internal confidence a model associates with the epistemic marker ``I think.” We calculate marker internal confidences (\mic s) for 13 models across 11 datasets.}
    \label{fig:fig1}
    \vspace{-5mm}
\end{figure}
While traditional uncertainty estimation methods \cite{huang2024surveyuncertaintyestimationllms, xia2025survey} predominantly rely on numerical representations of confidence (e.g., “25\% confident”), such approaches are limited by lack of alignment with natural communication and computational expense, especially if dependent on multiple model calls or auxiliary networks \cite{tian-etal-2023-just}.
Enabling models to use \textit{hedge phrases} or \textit{epistemic markers} to signal uncertainty (e.g., ``I think...'' or ``it is likely that...'') has therefore gained traction as a more effective alternative \cite{10.1145/3351095.3372852, 10.1145/3491102.3517791}. 
Hedging expressions are essential for effective communication \cite{willems, belem-etal-2024-perceptions} and can be integrated seamlessly into downstream settings to serve as a human-aligned medium for users to calibrate their reliance on model outputs.
Compared to numerical confidence, linguistic confidence expression is known \cite{Zimmer1983VerbalVN, BUDESCU1985391, wallsten1993preferences, 10.1145/3359206, dhami} to improve user judgments of LLM credibility, boost task accuracy in human-AI teaming, and incur minimal computational overhead. As such, recent work has increasingly focused on enabling LLMs to use natural language to \textit{faithfully} express their intrinsic uncertainty, which may be estimated via sampling consistency \cite{pmlr-v238-harsha-tanneru24a, gm, yona, metafaith}.\footnote{Note that use of a linguistic marker itself is low-overhead at deployment time; sampling consistency is an offline diagnostic used to estimate model confidence in a generation.}

Despite the importance of aligning LLMs’ internal and linguistically expressed confidence,
the consistency and reliability with which LLMs use epistemic markers to represent specific confidence levels remains poorly understood. 
Prior works \cite{gm, yona, ji, metafaith} have attempted to do so by comparing LLM and human recognition of epistemic markers, finding systematic deficiencies in LLMs' ability to faithfully express their (estimated) intrinsic confidence in words.
Yet these studies do not consider differences in human interpretations of markers across contexts, and rely on interpreting epistemic markers in a human fashion---it remains critically unclear whether models could be faithfully calibrated when interpreting markers according to their own use. 
In particular, even humans vary in individual-level interpretation and use of hedging language \cite{pennekamp2024variability}. A model might use ``perhaps" differently from humans, but still use it consistently whenever its own confidence is high. Thus, even if a model’s use of epistemic markers does not align with human intuition, such indicators can still be meaningful if the model applies a consistent mapping between utilized markers and internal confidence. We test whether this weaker, model-internal notion holds. This possibility of \textit{model-centric} faithful calibration is unexamined in prior work. 

To address this gap, we systematically examine whether epistemic markers emitted by LLMs reliably and stably reflect their intrinsic confidence across task settings. We formalize our study by first defining \textit{marker internal confidence} (\mic) (\S\ref{sec:mic}) as the average internal confidence of a model when it uses a given epistemic marker during a given task (Fig. \ref{fig:fig1}). This notably differs from prior work in that we measure the association between markers and intrinsic confidence, rather than with external factuality verdicts, and consider the impact of implicit contextual factors such as task domain, format, and difficulty.
Next, we propose a suite of 7 metrics (\S\ref{sec:metrics}) to assess the in- and out-of-distribution consistency of \mic s per model. We evaluate and analyze \mic s for 13 leading open-source and proprietary LLMs representing various sizes, families, and post-training procedures on 11 datasets.
Taken together, our results demonstrate that LLMs fail to consistently apply their own linguistic confidence framework. This provides critical evidence complementary to prior work \citep{metafaith} that LLMs are faithfully miscalibrated even under model-centric interpretation of marker meanings.

\section{Related Work} \label{sec:rw}
A plethora of approaches exist to numerically and linguistically assess the confidence of LLMs. 
We discuss such methods in greater detail in \S\ref{app:rw}.
In general, there is limited understanding of the ability for LLMs to consistently and reliably use epistemic markers to signal intrinsic uncertainty. Existing evaluations of the faithfulness of linguistic confidence expression by LLMs hinge on human perceptions of decisiveness \cite{yona, ji, metafaith, sft}, failing to consider (a) whether alignment is present if expressed confidence is defined according to a model’s own use, and (b) whether contextual features impact this alignment. 
We address these gaps by presenting a novel analysis framework and set of metrics specific to understanding the faithfulness and consistency with which LLMs use markers to express internal uncertainty in different contexts. 
This provides distinct insights from traditional studies of calibration, which are focused on alignment with factuality, and enables model-centric evaluation of the \textit{faithful calibration} of LLMs, complementary to existing studies of faithful linguistic confidence expression. To the best of our knowledge, we are the first to evaluate this possibility.

\section{Marker Internal Confidence} \label{sec:mic}

Our goal is to investigate when and to what extent models are able to consistently use specific epistemic markers\footnote{An epistemic marker is a word or phrase used to express ambiguity, caution, or indecisiveness about the content of a text \cite{lakoff, kranich, Juanchich}.
} to reflect specific internal confidence levels. To this end, we operationalize the internal confidence level of a model associated with an epistemic marker as the \textit{marker internal confidence}. 
Let $E$ be an epistemic marker, and suppose we have a dataset $D$ = $(D_{\text{train}}, D_{\text{test}})$ and a model $M$ which emits uncertainty-bearing responses $R_1,\ldots, R_N$ to the questions in the training set $D_{\text{train}} = \{Q_1, \ldots, Q_N\}$. 
We view each model response $R_i$ as a sequence of $L_i$ sentences $\{s_{i,l}\}_{l=1}^{L_i}$.
The marker internal confidence (\mic) of $E$ with respect to $M$ and $D$ is then computed as the average internal confidence of the model in the sentences in which $E$ appears. That is:
\begin{equation}
    \mic_{E,M,D} = \frac{1}{S}\sum_{i=1}^N \sum_{l=1}^{L_i} \mathds{1}\sbrace{E\in s_{i,l}} \conf(s_{i,l})  \label{eq:mic}
\end{equation}
where $\mathds{1}(\cdot)\in\{0,1\}$ is the indicator for whether sentence $s_{i,l}$ contains marker $E$, $\conf(s_{i,l})\in[0,1]$ is the intrinsic confidence of $M$ in sentence $s_{i,l}$, and $S:=\sum_{i=1}^N \sum_{l=1}^{L_i} \mathds{1}\sbrace{E\in s_{i,l}}$ is the total number of sentences containing $E$.\footnote{Following \citet{marconf}, we assume each sentence $s_{i,l}$ uses at most one marker. We demonstrate empirically that this requirement is not overly artificial in \S\ref{app:addl_verif}, by comparing the average number of hedges used per sentence with versus without such specification for representative models.
}

\paragraph{Quantifying Intrinsic Confidence.}
Following prior work \cite{metafaith, sft, ji, yona, manakul-etal-2023-selfcheckgpt, tian-etal-2023-just}, we quantify models' internal confidence by assessing consistency across sampled responses.
In particular, we adopt the procedure of \citet{metafaith}, as this methodology has been shown to address limitations\footnote{For example, assuming the same number and order of assertions across sampled responses.} of the earlier approach proposed by \citet{yona} while being provably robust (\S\ref{app:conf}) across diverse prompting strategies, models, and tasks.
Under this approach, given a text input $Q$ and response $R=\{s_{l}\}_{l=1}^{L}$, we sample $K$ additional responses $R_1,\ldots,R_K$ to $Q$ and use 
LLM-as-a-Judge
to assess whether each $s_l$ is contradicted by the sampled responses. The confidence $\conf(s_l)$ is then computed as the fraction of contradictory responses: $1 - \frac{1}{K} \sum_{k}[R_k\bot s_l].$
Details can be seen in \S\ref{app:conf}.

\paragraph{Evaluation Metrics.} \label{sec:metrics}
To systematically evaluate the consistency and stability of \mic s across models, datasets, and epistemic markers, we propose 7 model-level metrics which provide detailed insights into the distributions, generalizability, and confidence meanings of \mic s. Exact formulas and further details about the implementation and interpretation of these metrics can be found in \S\ref{app:metrics}:
\begin{itemize}[topsep=0pt, align=left, leftmargin=15pt, labelindent=1pt, listparindent=\parindent, labelwidth=0pt, itemindent=!, itemsep=0pt, parsep=0pt]
\item \imae: The \textit{in-domain average mean absolute error (MAE)} estimates how well a model's \mic s align with its actual intrinsic confidence on in-distribution test tasks (stability across similar contexts). 
It is computed by generating model responses on the test set $\dtest$ of each dataset, computing for each observed marker the absolute error between the \micx estimated on $\dtrain$ and the model's internal confidence in each sentence containing the marker, averaging across such sentences, averaging across markers, and finally averaging across datasets. We average over markers to bypass potential bias due to different relative marker frequencies across model responses per dataset; two variants which aggregate instead over sentences or samples are discussed in \S\ref{app:metrics}, with additional results in \S\ref{app:aggregation}. The \imaex ranges in value from 0 to 1, with lower \imaex representing better in-domain train-test transferability of \mic s.
\item \cmae: The \textit{cross-domain average MAE} assesses how well a model's \mic s generalize to align with intrinsic confidence on out-of-distribution test tasks. 
It is computed similarly to the \imae, but uses train and test sets from \textit{different} tasks and averages over pairs of distinct datasets instead of over singular datasets. Thus, it ranges from 0 to 1, with lower \cmaex indicating better cross-domain transferability of \mic s. The impact of aggregation granularity on \cmaex results is also studied in \S\ref{app:aggregation}.
\item \mcv: The \textit{marker-level average coefficient of variation (CV)} evaluates the consistency of a model's per-marker \micx values across datasets. It is computed as the CV of each shared\footnote{Shared markers are those which appear in a model's responses at least ten times per task. This threshold is determined sufficient based on empirical analysis (\S\ref{subsec:threshold}) and inspired by prior work \cite{marconf}.} marker's per-dataset \micx values, averaged \textit{across markers}.
The CV has no theoretical maximum, but a low \mcvx reflects stable cross-dataset \micx values per marker, suggesting consistent internal confidence is associated to each marker regardless of task setting.
\item \dcv: The \textit{dataset-level average coefficient of variation (CV)} measures the dispersion of a model's \micx values within each dataset. It is computed as the CV of all \mic s per dataset, averaged \textit{across datasets}. 
Similar to the \mcv, a low \dcvx indicates that marker values are highly concentrated, suggesting the model fails to meaningfully differentiate its use of epistemic markers in a generalized fashion.
\item \mrc: \textit{Marker rank correlation} measures the consistency of a model's \mic-based marker rankings across datasets. 
It assesses ranking of shared markers between pairs of datasets and is computed as the Fisher\footnote{We use the Fisher-transformed mean rather than the arithmetic mean when averaging correlation coefficients, as the latter is statistically unsound due to the bounded and nonlinear nature of correlation values \cite{7e2958c8-cf46-3edc-b197-57ee54882a19, Fisher014OT}. 
} average Spearman correlation of shared-marker \mic s across distinct dataset pairs. The \mrcx ranges from $-1$ to 1, with 1 ($-1$) indicating nearly identical (inverse) rankings of markers' associated intrinsic confidence, and 0 indicating minimal correlation.
\item \mac: The \textit{\mic-accuracy correlation} measures the cross-dataset association between a model's \micx values and accuracy. 
It is computed as the Fisher average across markers of the Pearson correlation between per-dataset \mic s and per-dataset accuracies (Spearman correlation results in \S\ref{app:corr}). 
The \macx ranges from $-1$ to 1. High \macx indicates the \mic s track dataset difficulty to reflect performance 
without encoding a differentiated internal confidence scale. 

\item \mcc: The \textit{\mic-\cmfgx correlation} measures the cross-dataset association between a model's \micx values and faithful calibration (FC) level, serving as the faithfulness-based analog to \macx. 
FC is the alignment between a model's internal and linguistically expressed confidence, estimated by the \cmfgx metric \cite{yona, metafaith, sft}, which quantifies internal confidence via sampling consistency and linguistic confidence via human-perceived decisiveness. The \mccx is computed by substituting \cmfgx for accuracy in the \macx computation (Spearman results in \S\ref{app:corr}).
It ranges from $-1$ to 1; high \mccx indicates \mic s are predictive of dataset-level FC, suggesting internal sensitivity to linguistic confidence expression.
\end{itemize}

\section{Experimental Setup} \label{sec:exps}

\paragraph{Models \& Datasets.} 

We compute and analyze \mic s for 13 leading open-source and proprietary models (\S\ref{app:models}) over 11 datasets (\S\ref{app:datasets}) spanning diverse content domains, task formats, and difficulty levels to ensure generality of our findings.

\paragraph{Prompts.} Standardized zero-shot task prompts are used to elicit model responses per dataset. Following prior work \cite{yona, metafaith}, in our experiments the task prompt is paired with a generic system prompt instructing the model to provide its answer succinctly while using at most one epistemic marker per sentence when uncertain. Additional results of using a metacognitive system prompt \cite{metafaith} to enhance the faithfulness of models' linguistic uncertainty expression are in \S\ref{subsec:sysprompt}. Full prompts are in \S\ref{app:prompts}. 

\paragraph{Marker Extraction.} \label{subsec:extraction} To compute \mic s, we first parse each model response into sentences\footnote{We use \texttt{spaCy} for sentence segmentation as it is a popular approach which achieves strong empirical performance on diverse text types. Segmentation quality was verified by the authors via manual review of 300 samples.} and prompt a strong LLM\footnote{We use Gemini-2.5-Flash-Lite as it is a capable and cost-efficient model. To verify the efficacy and validity of our extraction prompt, we manually annotate 300 samples and find a precision of 0.97 and recall of 1.0 against human annotations.} to extract and standardize epistemic markers per sentence. Sentences with more than one marker are discarded; we consider only markers which appear at least $T=10$ times per model. This threshold balances marker diversity against the need for sufficient samples per marker for reliable \micx estimation. Robustness to $T$ value is analyzed in \S\ref{subsec:threshold}. The use of no marker is considered as a special marker, \texttt{<no\_hedge>}.

\section{Results \& Analysis} \label{sec:results}

\begin{table*}[t]
\centering\footnotesize
\setlength{\tabcolsep}{5pt}
\caption{\textbf{Consistency and stability of marker internal confidences per model,} evaluated via metrics introduced in \S\ref{sec:metrics}. $\mae$ metrics range from 0 to 1, rank metrics from $-1$ to 1, and $\cv$ metrics start at 0 with no theoretical maximum.
In general, \mic s transfer reasonably well within a task, degrade across tasks, occupy a compressed value range, and show ranking consistency that is weaker once \texttt{<no\_hedge>} is removed.
}
\begin{tabular}{@{}lccccccc@{}}
\toprule
& \multicolumn{3}{c}{Marker Internal Confidence} & Density & \multicolumn{3}{c}{Rank}\\
\cmidrule(lr){2-4} \cmidrule(lr){5-5} \cmidrule(lr){6-8}
Model & \imaex $\downarrow$ & \cmaex$\downarrow$ & \mcvx$\downarrow$ & \dcvx & \mrcx$\uparrow$ & \macx & \mccx \\ \midrule
Gemini-2.5-Flash & 0.11 & 0.22 & 0.33 & 0.22 & \textbf{0.86} & 0.47 & 0.35 \\ 
Gemini-3-Flash & 0.09 & 0.13 & \textbf{0.11} & 0.08 & 0.76 & 0.26 & -0.66 \\ 
Gemini-3.1-Pro & \textbf{0.08} & \textbf{0.12} & \textbf{0.11} & 0.08 & 0.32 & 0.07 & -0.49 \\ \midrule
GPT-5-Nano & 0.15 & 0.26 & 0.32 & 0.27 & \textbf{0.90} & 0.62 & -0.40 \\ 
GPT-5-Mini & \textbf{0.11} & \textbf{0.19} & 0.29 & 0.28 & 0.70 & 0.60 & -0.02 \\
GPT-5 & \textbf{0.11} & 0.26 & \textbf{0.18} & 0.16 & 0.83 & 0.63 & -0.21 \\ \midrule
Qwen3-0.6B & 0.14 & 0.26 & 0.27 & 0.24 & 0.69 & 0.70 & 0.68 \\ 
Qwen3-1.7B & 0.12 & 0.21 & 0.27 & 0.24 & 0.77 & 0.51 & 0.15 \\ 
Qwen3-4B & \textbf{0.10} & 0.20 & 0.22 & 0.20 & 0.79 & 0.61 & 0.23 \\ 
Qwen3-8B & \textbf{0.10} & 0.15 & 0.21 & 0.18 & 0.85 & 0.65 & -0.24 \\ 
Qwen3-32B & 0.11 & \textbf{0.13} & \textbf{0.16} & 0.17 & \textbf{0.88} & 0.66 & -0.11 \\ \midrule
Llama3.1-8B-Instruct & 0.16 & 0.29 & 0.40 & 0.29 & 0.46 & 0.69 & 0.30 \\ 
Llama3.3-70B-instruct & \textbf{0.14} & \textbf{0.21} & \textbf{0.21} & 0.19 & \textbf{0.68} & 0.34 & -0.45 \\ \bottomrule
\end{tabular}
\label{tab:main}
\vspace{-4mm}
\end{table*}

\subsection{Main Results} \label{subsec:mainresults}
We report\footnote{As \imaex and \cmaex are computed as averages of means, to capture variability we additionally provide the pooled standard deviation per score in \S\ref{app:aggregation}.} results in Table \ref{tab:main} and 
observe:

\textbf{The stability with which models use epistemic markers to reflect internal confidence is sensitive to context and distribution shifts}. While \mic s are fairly stable within the same task setting, indicated by low \imaex values, cross-dataset robustness is weaker. \cmaex is consistently higher than \imae, and moderate \mcvx values indicate that the internal confidence associated with each marker fluctuates across datasets, with models struggling to generalize their use of epistemic markers to reflect specific internal confidence across tasks.

\textbf{Models exhibit limited ability to discriminate markers by internal confidence.} Across models, \dcvx values are quite low (0.17--0.29), demonstrating that \mic s are highly concentrated. This is not attributable to a limited range of observed internal confidences: per-task confidence ranges generally span 0--1, and the Pearson and Spearman correlations between \dcvx and average confidence range across datasets are near-0 or insignificant. This evidences that models struggle to convey internal confidence faithfully even when interpreting epistemic markers on their own terms, providing complementary yet critical evidence prior work \cite{metafaith, yona} to show that models are holistically unfaithfully calibrated.

\begin{wrapfigure}{r}{0.45\textwidth}
    \vspace{-14pt} %
    \centering
    \includegraphics[width=\linewidth]{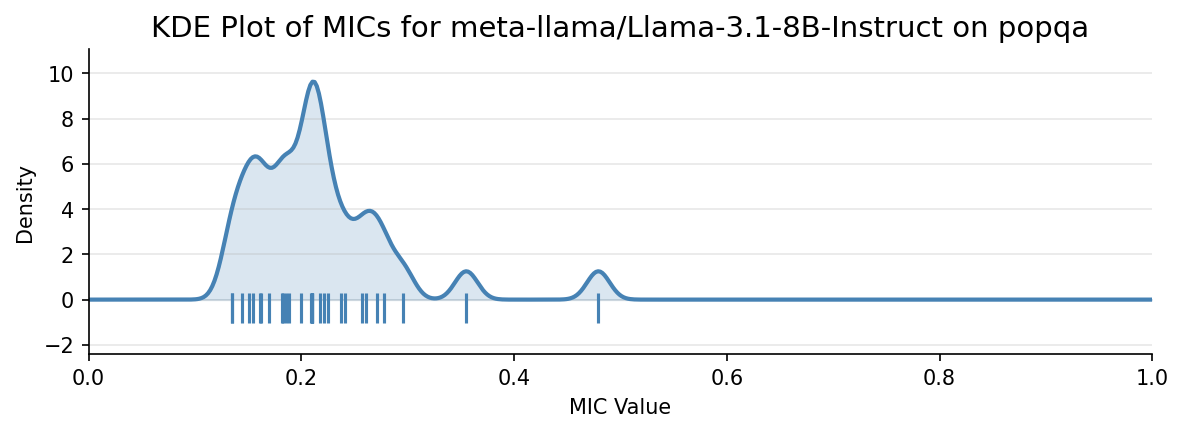}\\[4pt]
    \includegraphics[width=\linewidth]{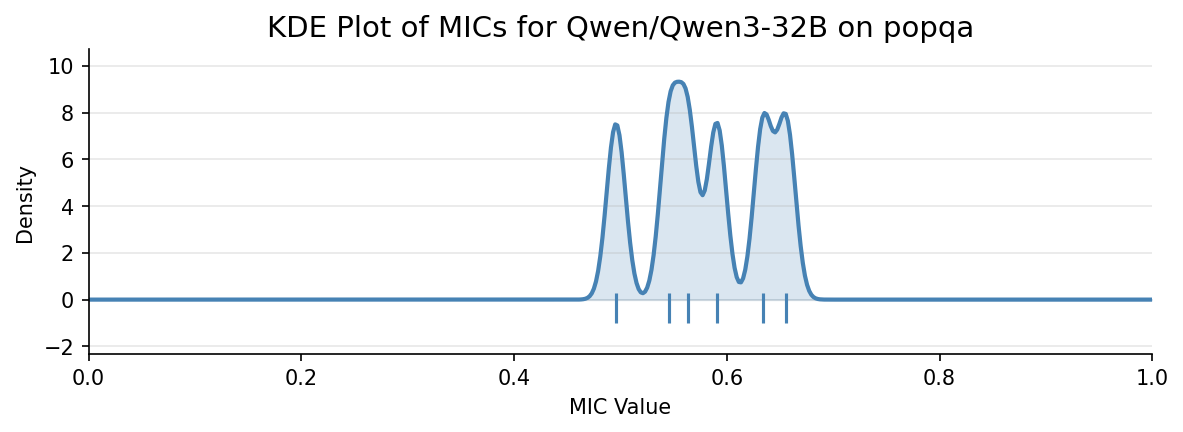}
    \caption{KDE plots of \micx values on PopQA.}\label{fig:kde}
    \vspace{-20pt} %
\end{wrapfigure}
\textbf{Despite limited marker discriminability, models encode multiple 
coarse 
uncertainty levels.} 
Low \dcvx values 
alone do not reveal whether markers form a single undifferentiated cluster or a small number of stable but closely spaced uncertainty bands. To investigate this, we plot the kernel density estimate (KDE) of \micx values for each model and dataset, which reveals the underlying distributional structure without requiring a priori cluster assumptions. Representative plots for Llama3.1-8B-Instruct and Qwen3-32B on the dataset PopQA are shown in Fig. \ref{fig:kde} (plots for all datasets in \S\ref{app:visualizations}). Consistent with the low \dcvx values, \micx distributions are concentrated within a narrow region of the confidence space. 
However, internal structure varies across models: smaller models tend to exhibit unstable, imbalanced distributions with one dominant peak and minor secondary peaks, while larger models show clearer multimodal structure with several peaks of comparable prominence. This suggests larger models can encode a greater number of stable uncertainty bands within the compressed confidence range.
Thus, although fine-grained discrimination among markers is limited, models nevertheless encode distinct uncertainty levels through their marker use.

\begin{table*}[t]
\centering\footnotesize
\setlength{\tabcolsep}{5pt}
\caption{Excerpted analysis of \mic s when the special marker \texttt{<no\_hedge>} is excluded. $\Delta \mrc$ indicates the change in \mrcx after the exclusion, in comparison to original \mrcx scores in Table \ref{tab:main}. Full results are in \S\ref{app:nohedge}.}
\begin{tabular}{@{}lccccc >{\columncolor{blue!10}}ccc@{}}
\toprule
& \multicolumn{3}{c}{Marker Internal Confidence} & Density & \multicolumn{3}{c}{Rank}\\
\cmidrule(lr){2-4} \cmidrule(lr){5-5} \cmidrule(lr){6-9}
Model & \imaex $\downarrow$ & \cmaex$\downarrow$ & \mcvx$\downarrow$ & \dcvx & \mrcx$\uparrow$ & $\Delta \mrc$ & \macx & \mccx \\ \midrule
Gemini-2.5-Flash & 0.11 & 0.22 & 0.34 & 0.22 & 0.39 & $-0.47$ & 0.52 & 0.37 \\ 
Llama3.1-8B-Instruct & 0.16 & 0.29 & 0.40 & 0.29 & 0.18& $-0.28$ & 0.61 & 0.33 \\ 
Qwen3-1.7B & 0.12 & 0.21 & 0.26 & 0.24 & 0.78& $+0.01$ & 0.60 & 0.17 \\ 
Qwen3-4B & 0.10 & 0.20 & 0.22 & 0.18 & 0.47& $-0.32$ & 0.53 & 0.20 \\ 
Qwen3-32B & 0.11 & 0.17 & 0.21 & 0.20 & 0.27& $-0.61$ & 0.59 & -0.05 \\ \midrule
\end{tabular}
\label{tab:nohedge}
\vspace{-6mm}
\end{table*}

\textbf{Models maintain consistent rank ordering of epistemic markers across tasks.} Most models achieve $\mrc\geq 0.68$, indicating that although they struggle to differentiate the confidence meaning of individual markers, their relative ordering is fairly well-preserved across distributions. Models may possess implicit understanding of internal confidence levels despite limited ability to use markers in a meaningful fashion \citep{metafaith}. However, increased model size does not always help, especially for proprietary LLMs.

\textbf{Marker ranking consistency is partly driven by the absence of hedging.} To determine whether observed trends truly reflect models' ability to rank and differentiate among individual markers,
we repeat our analysis while excluding the special marker \texttt{<no\_hedge>}. Abbreviated results are reported in Table \ref{tab:nohedge}, with full results in \S\ref{app:nohedge}. We observe that the $\mae$- and $\cv$-based metrics remain largely stable, indicating reliable results regarding discriminability and consistency of \mic s attributable to true graded differences (albeit slight) among different markers' associated internal confidences. Associations between \micx and accuracy (\mac) or faithful calibration (\mcc) are likewise generally preserved. However, \mrcx declines notably for most models (see blue column), suggesting that \texttt{<no\_hedge>} acts as a strong anchor in marker rankings, and that genuine ranking consistency among markers is weak. Overall, this indicates that while models can form a somewhat meaningful graded lexicon of epistemic markers, the apparent consistency of marker rankings is largely mediated by the distinction between hedging vs. no hedging, rather than by reliable encoding of marker semantics.

\textbf{Increased model size can improve stability and meaningfulness of \mic s, but not discriminability.} 
Within each model family, larger and stronger models consistently achieve smaller \imae, \cmae, and \mcvx and often higher \mrc, demonstrating improved ability to consistently associate and stably order markers by internal confidence across tasks. This is supported by moderate correlations between cross-dataset average accuracy (representative of model capability) and \cmae, \mcv, \dcv, and \mrcx (Pearson: $-0.53, -0.44, -0.61$, 0.33; Spearman: $0.51$, $-0.45$, $-0.58, 0.25$; $p<0.05$). At the same time, lower \dcvx for larger models indicates reduced differentiation of markers based on internal confidence. Finally, model size appears to have limited impact on the association between \micx values and performance (\mac) or faithful calibration (\mcc).

\textbf{\mic s track accuracy but not faithful calibration.} For several models, \macx exceeds 0.6, indicating \mic s are moderately positively associated with accuracy. This shows that \micx is partly driven by dataset difficulty, consistent with earlier-observed fragility under distribution shifts. 
Interestingly, higher \macx suggests that models could achieve good faithful calibration by using linguistic confidence expressions that reflect task accuracy or difficulty as a proxy for internal confidence; this would indicate that faithful and factuality-aligned calibration can indeed be achieved simultaneously, in opposition to prior findings by \citet{metafaith} for select models and calibration approaches. Moreover, if models can harness internal signals regarding task difficulty or performance to faithfully represent their intrinsic confidence, this would signal that LLMs indeed have the ability to be aware of and use metacognitive information, which is critical to enhance human-AI interaction and self-driven learning \cite{steyvers2025metacognition}.

\begin{figure*}[t]
    \centering
    \begin{minipage}[t]{0.48\textwidth}
        \centering
        \includegraphics[width=0.9\linewidth]{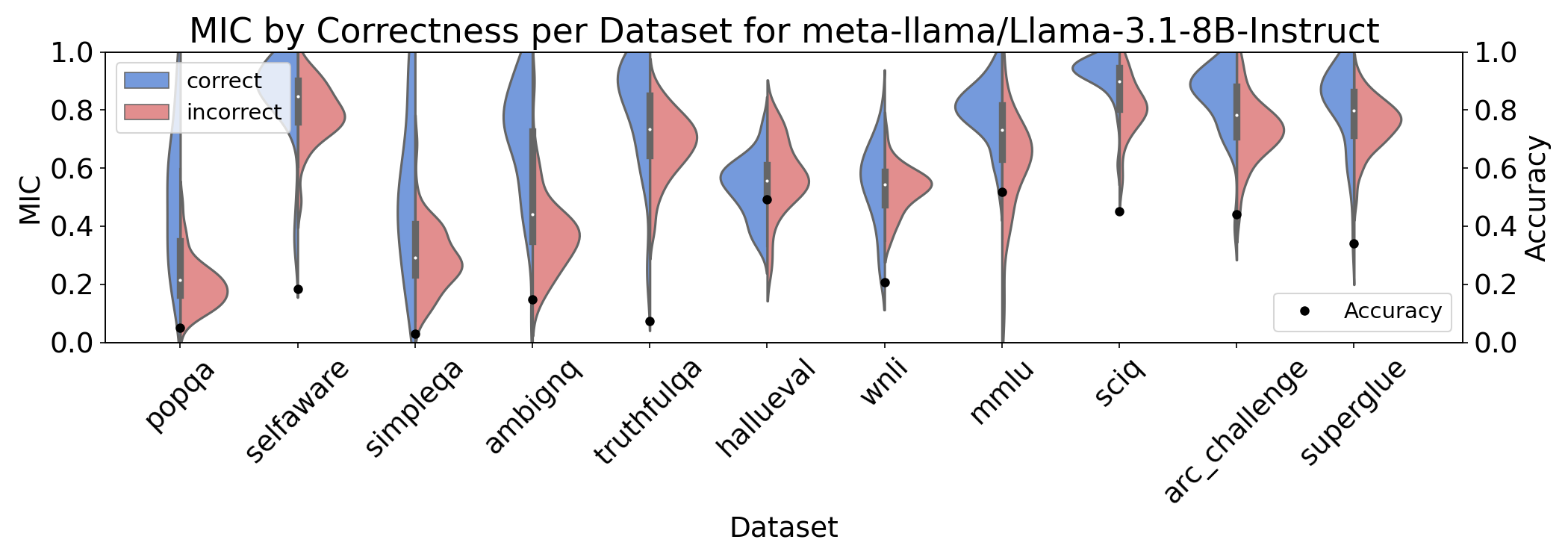}
    \end{minipage}
    \hfill
    \begin{minipage}[t]{0.48\textwidth}
        \centering
        \includegraphics[width=0.9\linewidth]{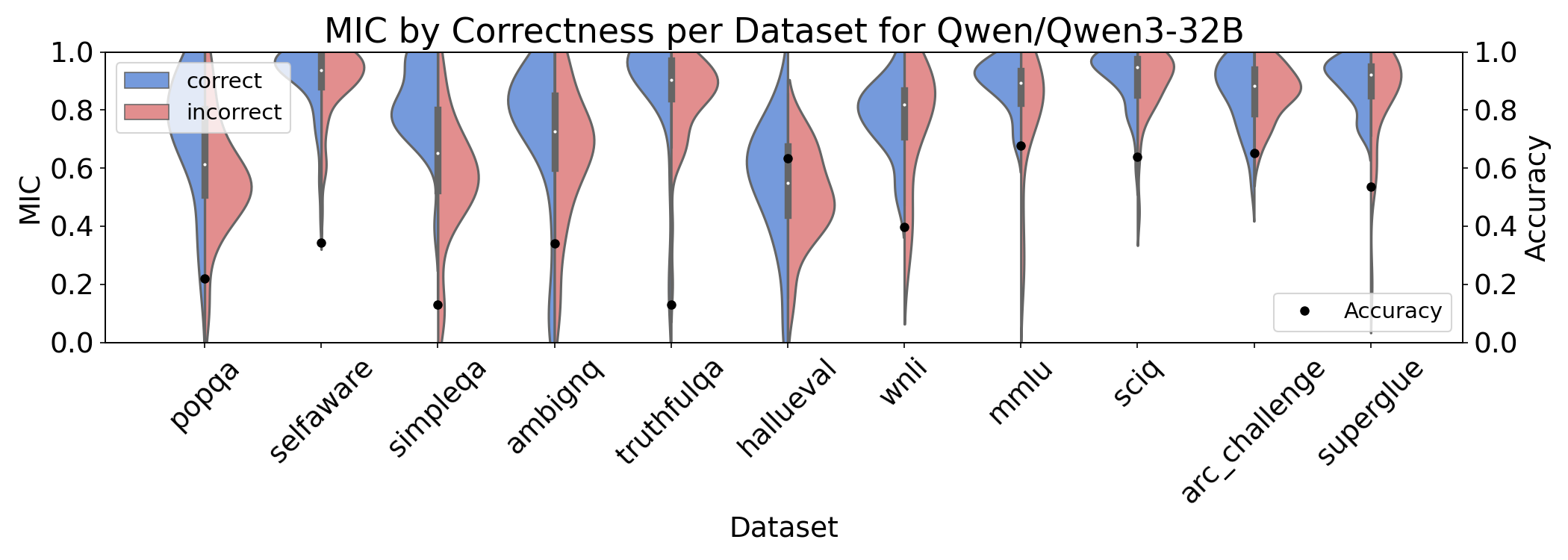}
    \end{minipage}
    \caption{\textbf{Representative violin plots of models' \micx densities across datasets, stratified by correctness.} Dataset-level accuracy is indicated by black points (values along the second $y$-axis). 
    }\label{fig:ic}
\end{figure*}

\begin{figure*}[t]
    \centering
    \begin{minipage}[t]{0.48\textwidth}
        \centering
        \includegraphics[width=0.9\linewidth]{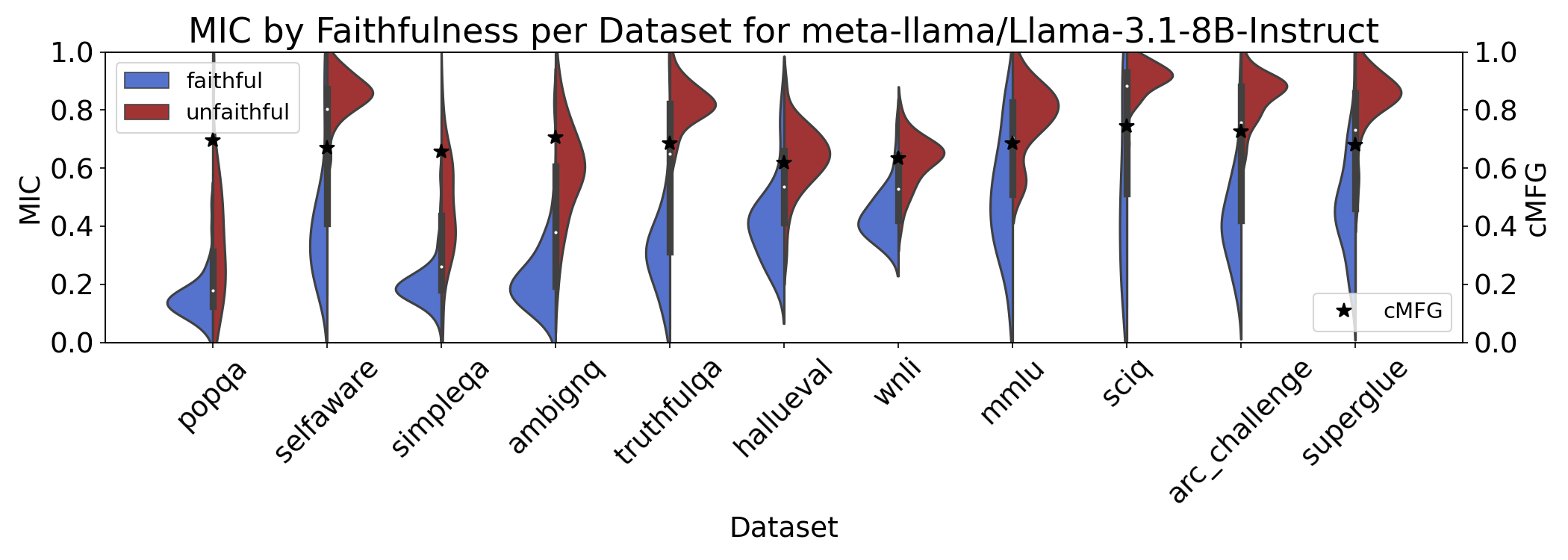}
    \end{minipage}
    \hfill
    \begin{minipage}[t]{0.48\textwidth}
        \centering
        \includegraphics[width=0.9\linewidth]{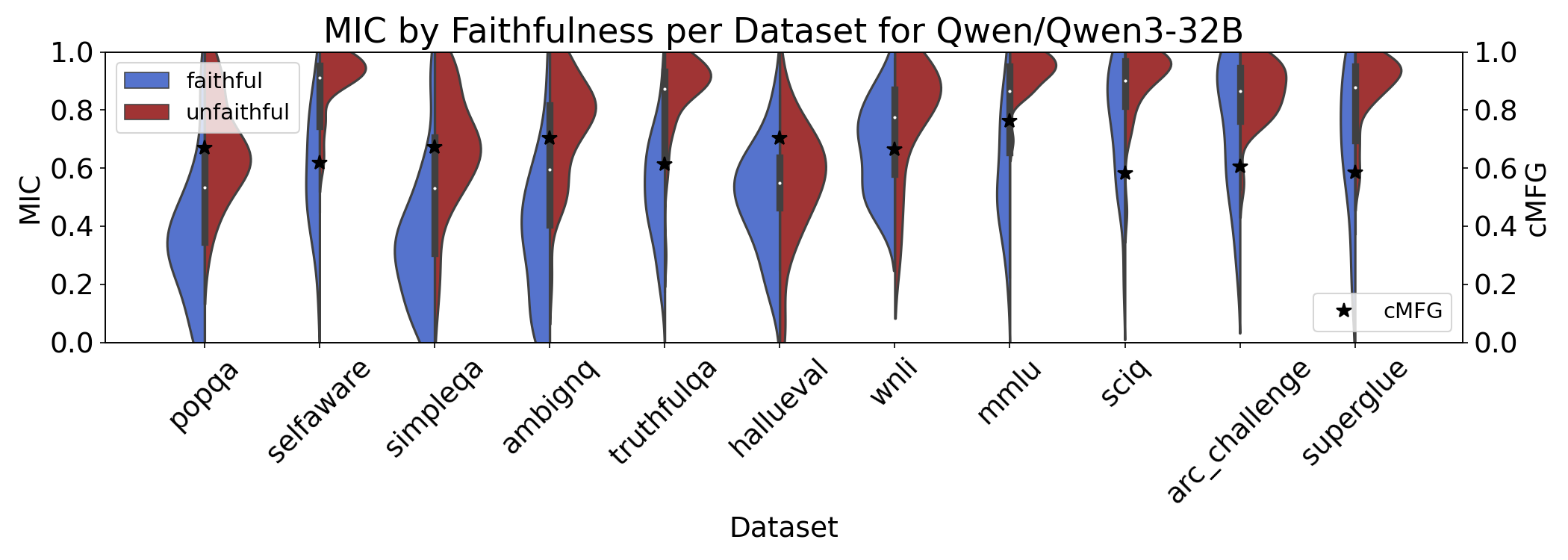}
    \end{minipage}
    \caption{\textbf{Representative violin plots of models' \micx densities across datasets, stratified by faithful calibration level.} 
    Dataset-level faithful calibration is indicated by black stars (values along the second $y$-axis). 
    }\label{fig:uf}
\end{figure*}
\begin{figure*}[h!]
    \centering
    \begin{minipage}[t]{0.45\textwidth}
        \centering
        \includegraphics[width=0.9\linewidth]{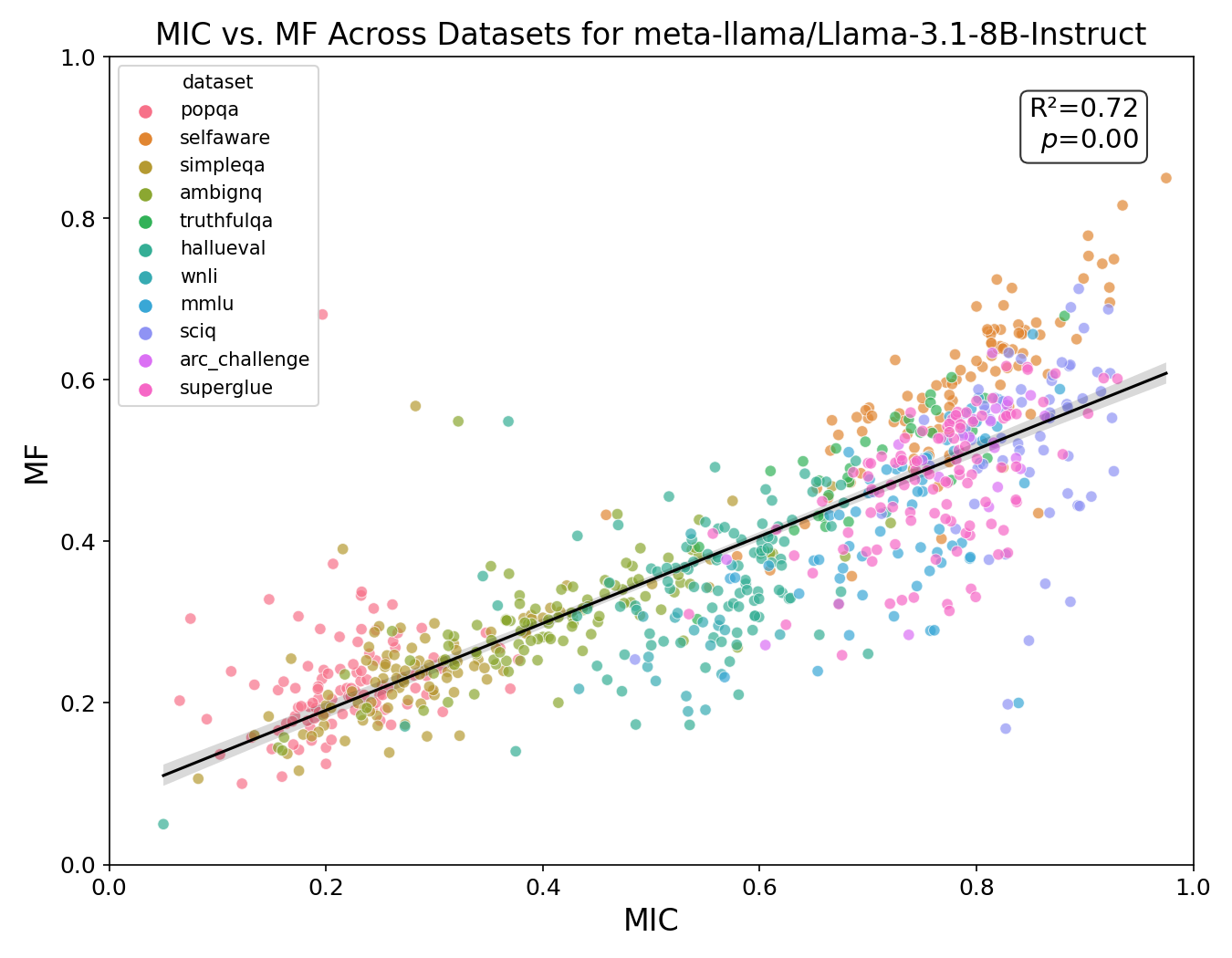}
    \end{minipage}
    \hfill
    \begin{minipage}[t]{0.45\textwidth}
        \centering
        \includegraphics[width=0.9\linewidth]{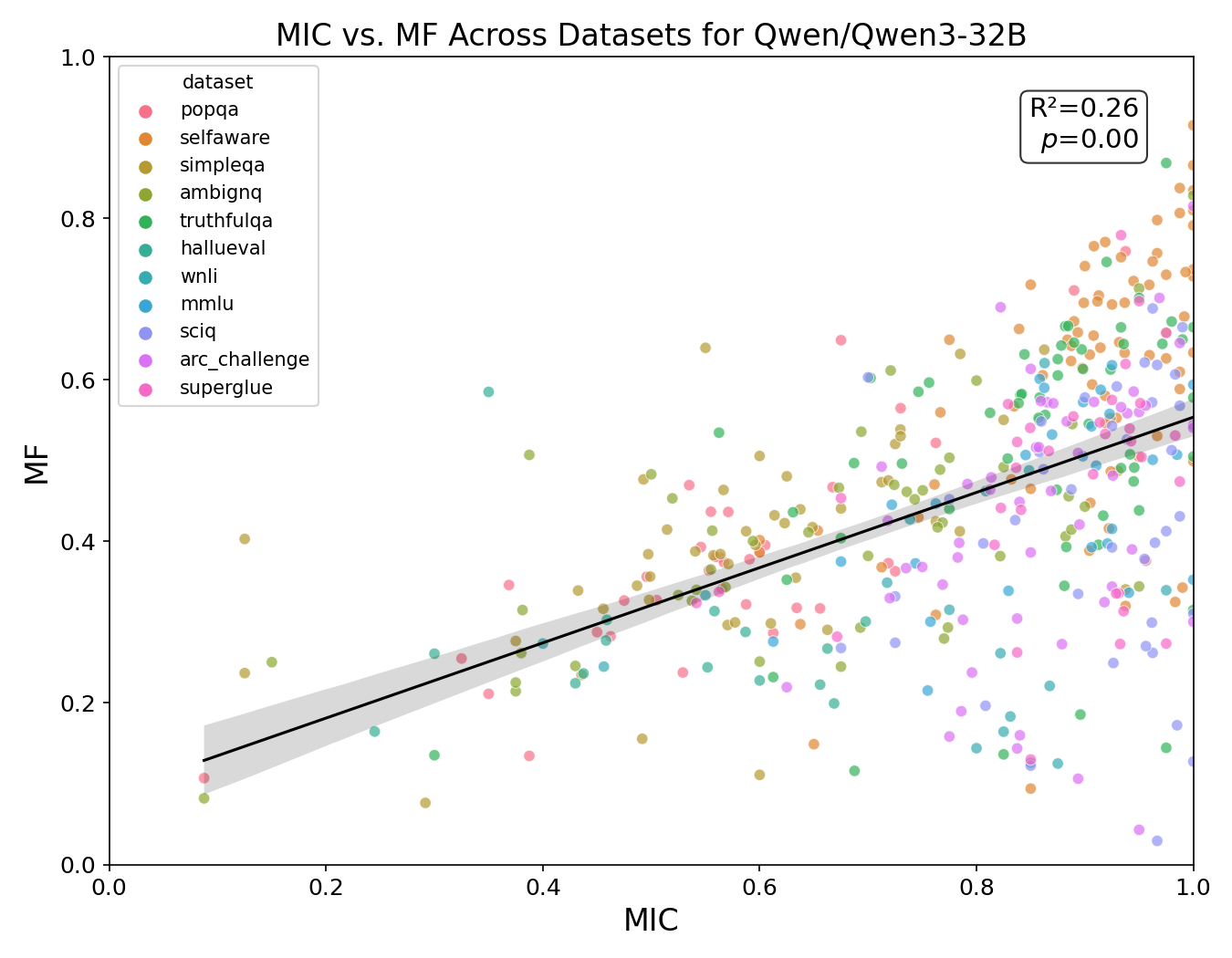}
    \end{minipage}
    \caption{\textbf{Representative plots of models' \mic s in relation to per-marker average absolute difference between internal confidence and human-interpreted linguistic decisiveness (\mf).} We observe positive trends between internal confidence and human-model decisiveness divergence per marker, suggesting high-\micx markers are a primary driver of faithful miscalibration in LLMs when we interpret markers according to human perception.}\label{fig:mf}
\vspace{-4mm}
\end{figure*}

On the other hand, \mccx results are mixed: smaller models 
achieve $\mcc\geq 0.15$, while larger models’ \mccx is $\leq0$. This indicates \mic s do not reliably reflect models’ ability to use markers in a human-aligned fashion, providing further evidence for divergence between human and LLM interpretation of linguistic uncertainty \cite{tang2026evaluation}.

We corroborate these findings by plotting \micx density stratified by correctness and faithful calibration level per model (representative plots for Llama3.1-8B-Instruct and Qwen3-32B in Fig.s \ref{fig:ic} and \ref{fig:uf}): \micx mass is generally lower for incorrect samples and higher for correct ones, suggesting models can use epistemic markers in association with correctness. 
In contrast, low-faithfulness samples\footnote{In line with prior work, we consider samples whose faithfulness score \cite{yona} is $\geq 0.75$ or $<0.75$ as relatively faithful or unfaithful, respectively.} 
concentrate at high \micx values (0.5--1.0), while more faithful samples exhibit lower, more distributed \mic s across nearly all datasets. This suggests high-\micx markers may be used automatically regardless of intrinsic uncertainty, 
while low-\micx markers reflect more genuine hedging. This is further supported by plots of \micx against the per-marker average absolute difference between human-interpreted decisiveness and internal confidence (\mf), shown in a representative fashion for Llama3.1-8B-Instruct and Qwen3-32B in Fig. \ref{fig:mf}, revealing positive per-model correlations between internal confidence and human-model decisiveness divergence per marker.
Taken together, our findings propose high-\micx markers as the primary driver of faithful calibration failures, helping to explain gaps in prior analysis \cite{metafaith} and providing a more coherent account of systematic limitations of LLMs in faithfully expressing intrinsic uncertainty.

\subsection{Impact of Marker Threshold} \label{subsec:threshold}

\begin{table*}[t]
\centering\footnotesize
\setlength{\tabcolsep}{7.5pt}
\caption{\textbf{Comparative impact of marker filtering threshold} on \micx analysis results for a representative sample of models. Main findings ($T=10$) are preserved regardless of threshold value.}
\begin{tabular}{@{}lcccc|cccc|cccc@{}}
\toprule
& \multicolumn{4}{c}{\mcvx $\downarrow$} & \multicolumn{4}{c}{\dcvx} & \multicolumn{4}{c}{\mrcx$\uparrow$}\\
\cmidrule(lr){2-5} \cmidrule(lr){6-9}\cmidrule(lr){10-13}
Threshold & $10$  &$20$  &$50$  &$100$  &10  &20  &50  &100 &10 &20 &50  &100 \\\midrule 
Gemini-2.5-Flash & 0.33 & 0.30 & 0.31 & 0.33 & 0.22 & 0.22 & 0.26 & 0.27 & 0.86 & 0.80 & 0.79 & 0.83 \\ 
Llama3.1-8B-Instruct & 0.40 & 0.37 & 0.38 & 0.36 & 0.29 & 0.23 & 0.18 & 0.25 & 0.46 & 0.51 & 0.43 & 0.47 \\ 
Qwen3-1.7B & 0.27 & 0.28 & 0.30 & 0.25 & 0.24 & 0.26 & 0.24 & 0.25 & 0.77 & 0.74 & 0.77 & 0.76 \\ 
Qwen3-4B & 0.22 & 0.21 & 0.21 & 0.22 & 0.20 & 0.19 & 0.19 & 0.21 & 0.79 & 0.77 & 0.76 & 0.75 \\ 
Qwen3-32B & 0.16 & 0.15 & 0.17 & 0.17 & 0.17 & 0.20 & 0.16 & 0.22 & 0.88 & 0.89 & 0.81 & 0.82 \\ \bottomrule
\end{tabular}
\label{tab:threshold}
\end{table*}
\begin{table*}[t]
\centering\footnotesize
\setlength{\tabcolsep}{7.5pt}
\caption{\textbf{Comparative impact of system prompt} on results for a representative sample of models. We compare the generic system prompt (Fig. \ref{fig:sys2}, \S\ref{app:prompts}) used in main experiments against a metacognitive system prompt (Fig. \ref{fig:sys3}, \S\ref{app:prompts}) known to improve human-aligned faithful uncertainty expression \cite{metafaith}.
}
\begin{tabular}{@{}lccccccc@{}}

\toprule
& \multicolumn{3}{c}{Marker Internal Confidence} & Density & \multicolumn{3}{c}{Rank}\\
\cmidrule(lr){2-4} \cmidrule(lr){5-5} \cmidrule(lr){6-8}

Model / System Prompt & \imaex $\downarrow$ & \cmaex $\downarrow$ & \mcvx $\downarrow$ & \dcvx & \mrcx$\uparrow$ & \macx & \mccx \\ \midrule
Gemini-2.5-Flash & 0.11 & 0.22 & 0.33 & 0.22 & 0.86 & 0.47 & 0.35 \\ 
+Metacognitive Prompt & 0.12 & 0.24 & 0.34 & 0.15 & 0.81 & 0.43 & 0.29 \\ \midrule
Llama3.1-8B-Instruct & 0.16 & 0.29 & 0.40 & 0.29 & 0.46 & 0.69 & 0.30 \\ 
+Metacognitive Prompt & 0.20 &  0.29 &  0.38 &  0.13 &  0.46 &  0.66 &  0.09 \\ \midrule
Qwen3-8B & 0.10 & 0.15 & 0.21 & 0.18 & 0.85 & 0.65 & -0.24 \\ 
+Metacognitive Prompt & 0.13 &  0.18 &  0.19 &  0.11 &  0.90 & 0.67 &  -0.41 \\ \midrule
Qwen3-32B & 0.11 & 0.13 & 0.16 & 0.17 & 0.88 & 0.66 & -0.11 \\ 
+Metacognitive Prompt & 0.13 &  0.16 &  0.22 &  0.12 &  0.79 &  0.61 &  -0.32 \\ \bottomrule
\end{tabular}
\label{tab:sysprompt}
\vspace{-4mm}
\end{table*}

The main experiments
consider only those markers which appear at least $T=10$ times per model per task. 
To demonstrate the robustness of our conclusions to the selection of $T$, we additionally investigate the use of $T=20, 50, 100$, which represent increased aggregate reliability of \micx estimates.\footnote{
These thresholds are determined in a principled fashion to span a graded range: $T=100$ represents the 30th percentile of marker frequency across all models, markers, and tasks; intermediate $T=20, 50$ approximately double their respective preceding frequencies. We empirically verify these thresholds are appropriate (e.g., to ensure $T=100$ is not too high) by computing the mean and standard deviation of $S$ across markers and tasks for each $T$. See \S\ref{app:addl_verif} for details.} 
Results for representative models are reported in Table \ref{tab:threshold}. As shown, even when the marker filtering threshold is increased to 100, the trends observed in \S\ref{subsec:mainresults} are persistent: 
\mcvx and \dcvx values remain low to moderate, and \mrcx values remain high, 
suggesting good ranking consistency but weak discriminability 
and limited robustness of \mic s to distribution shifts.
Larger models exhibit slightly better \micx stability and consistency but not necessarily discriminability, signaling model capability is important. 
Overall, this verifies the generalizability of our findings
and the associated implications for LLM uncertainty communication: models' shortcomings in reliable, meaningful marker use extend even to frequently occurring markers.

\subsection{Impact of System Prompt} \label{subsec:sysprompt}
Recent work \citep{metafaith} shows that models' ability to faithfully express their uncertainty by using linguistic markers in human-aligned fashion can be improved by adding \textit{metacognitive framing} to system prompts.
Thus, we observe the impact of such prompts on models' \mic s. 
Concretely, metacognitive prompting entails instructing a model that it has good \textit{metacognitive sensitivity} and privileged access to internal uncertainty signals.
We specifically use the prompt shown in Fig. \ref{fig:sys3} (\S\ref{app:prompts}) for a representative selection of LMs. Results are presented in Table \ref{tab:sysprompt}, which includes our main results as a baseline.
While the consistency and ranking of \mic s within and across tasks is largely unaffected by prompt strategy (see \imae, \cmae, \mcv, and \mrcx columns), discriminability worsens, reflected by consistently lower \dcvx with the metacognitive prompt. The association between \mic s and accuracy (\mac) remains stable, while the association with FC level (\mcc) is consistently lower under the metacognitive prompt. Thus, while metacognitive prompting can improve models' human-aligned faithful uncertainty expression, \mic s fail to track this change, suggesting a decoupling between internal confidence structures and surface-level uncertainty.

\section{Conclusion} \label{sec:conc}
In this work, we 
systematically examined
whether epistemic markers emitted by LLMs consistently and stably reflect their intrinsic confidence. We operationalized \textit{marker internal confidence} (\mic) as the internal confidence level an LLM associates with a specific marker in a given task setting, and introduced a suite of 7 metrics to evaluate the in- and out-of-distribution reliability of \mic s. Comprehensive experimentation and analysis revealed that modern LLMs struggle to reliably differentiate markers by internal confidence and habitually encode \mic s sensitive to distribution shifts and task difficulty.
At the same time, models somewhat preserved consistent marker rankings across tasks, and increased model size improved \micx stability but not discriminability. We also identified systemic limitations in models’ ability to faithfully express their intrinsic uncertainty in words, even when linguistic decisiveness is evaluated according to their own marker use. 
This shows that LLMs struggle to consistently apply their own linguistic confidence framework, rounding out prior work on faithful calibration of LLMs.
This underscores an alignment gap and the need to ground LLMs' epistemic marker use in more stable and meaningful internal confidence representations.

\section*{Limitations} \label{sec:lims}

Our work aims to establish an understanding of the reliability with which models use epistemic markers to reflect specific internal confidence levels. To this end, we focus on tasks in which models are expected to express a wide range of confidence values in their final answer, irrespective of the process used to obtain that response. Yet human language is complex and diverse, able to communicate rich connotations of uncertainty especially during the problem-solving process. It is therefore of interest to extend the present study to tasks involving complex reasoning and other types of long-form output—settings in which communicated uncertainty may not be fully captured by simply considering epistemic markers. Investigating use of multiple epistemic markers per sentence and analyzing \mic s at finer granularities presents another natural avenue for future work.

The impact of experimental factors such as temperature, prompting strategy, and post-training procedure (e.g., SFT vs. RLHF) on \micx stability also warrants investigation. Our experiments focus on RLHF-trained models as they possess strong instruction-following capabilities and represent the standard for evaluation of faithful uncertainty expression by LLMs \cite{metafaith}. While we show in \S\ref{subsec:sysprompt} that metacognitive prompting, known to improve human-aligned faithful calibration \cite{metafaith}, can somewhat improve \micx results, whether alternative strategies (e.g., use of few-shot examples derived using a model’s own \mic s) can fundamentally impact \micx consistency and stability is another interesting direction.

Finally, human epistemic marker use is known to vary significantly across cultures, languages, and contexts \citep{Lauwereyns_2002, YAGIZ2014260, socsci7040070, MURDUENAS2021103131}, so expanding our study to other cultural settings is another worthwhile path. Developing methods to improve the stability and consistency of models’ epistemic marker use across tasks
is another important challenge.

\bibliographystyle{plainnat}

\bibliography{anthology, custom, custom2}

\appendix
\appendix

\section{Related Work}\label{app:rw}

\paragraph{Confidence Estimation in LLMs.} There exist many methods to estimate the confidence level of LLMs \cite{huang2024surveyuncertaintyestimationllms, xia2025survey}. Such approaches can be broadly categorized as either black-box or white-box, depending on whether access to models’ internal states is required. Black-box methods assume access only to model outputs and include estimation based on sampling variability \cite{manakul-etal-2023-selfcheckgpt, becker2024cyclesthoughtmeasuringllm, chen-mueller-2024-quantifying, kaur2024addressing, xiong2024can}, semantic consistency \cite{meister-etal-2022-high, kuhn2023semantic,  grewal2024improvinguncertaintyquantificationlarge, nikitin2024kernellanguageentropyfinegrained}, direct prompting and verbalization \cite{cape, tian-etal-2023-just, hou2024decomposinguncertaintylargelanguage, yadkori2024believebelievellm, yang2024verbalizedconfidencescoresllms, zhao-etal-2024-fact}, or use of auxiliary predictive models \cite{shrivastava2023llamasknowgptsdont, shen2024thermometeruniversalcalibrationlarge}. In contrast, white-box methods leverage internal states of LLMs to estimate confidence by examining token probabilities \cite{duan-etal-2024-shifting, kadavath2022languagemodelsmostlyknow, Huang_2025}, probing internal representations \cite{azaria-mitchell-2023-internal, burns2024discoveringlatentknowledgelanguage, ji}, or training with uncertainty-augmented data \cite{lin2022teaching, chaudhry2024finetuninglanguagemodelsemit, lacie, zhang2024rtuninginstructinglargelanguage}, among others. Despite the efficacy of such approaches, they primarily focus on numerical confidence estimation and do not consider linguistic uncertainty expression, constitutionally ignoring the critical role of linguistic assertiveness in communicating model uncertainty to downstream users \cite{10.1145/3630106.3658941}. In contrast, use of epistemic markers to signal uncertainty yields significantly more expressivity and allows models to directly embed uncertainty into their outputs in a human-aligned fashion. Since it is generally unknown whether models have the ability to consistently and reliably do so, we tackle this gap by systematically analyzing models’ use of linguistic patterns to signal intrinsic confidence.

\paragraph{Linguistic Uncertainty in LLMs.} As natural language is the primary interface for human-LLM interaction, another line of work \cite{band2024linguisticcalibrationlongformgenerations, xiong2024can, yang2024alignmenthonesty, jiang2025conformallinguisticcalibrationtradingoff, tao2025can, wang2025calibrating, tang2026evaluation} has pursued the integration of linguistic uncertainty into model outputs through self-verbalization or mapping of numerical confidence scores to uncertainty phrases (e.g., “high confidence”). However, such works focus on aligning expressed confidence with accuracy as opposed to the true internal confidence of the model. Moreover, they often simplify the space of linguistic markers used \cite{zhou-etal-2024-relying}, failing to account for the plurality of human linguistic uncertainty expressions, and face challenges with generalizability. For example, \citet{mielke-etal-2022-reducing} utilize a limited scoring scale to measure confidence and assertiveness, while \citet{lin2022teaching} depend on computationally expensive, domain-specific training and does not enable zero-shot confidence verbalization. \citet{zhang-etal-2024-dont-go} further finds that verbalized confidences tend to concentrate in restricted ranges, leading to constrained efficacy in real-world settings.

\paragraph{LLM Use of Epistemic Markers.} More recently, the reliability and calibration with which LLMs can use epistemic markers to signal confidence has gained traction \cite{tang2026evaluation, lee2025llm, belem-etal-2024-perceptions, zhou-etal-2024-relying, zhou-etal-2023-navigating}.
For example, \citet{marconf} study the consistency with which LLMs use epistemic markers to reflect specific accuracy levels within and across tasks. However, their focus is on the ability for models to distinguish output correctness levels via linguistic confidence, which diverges from the aim of faithful linguistic expression of LLMs’ intrinsic confidence. It is generally well-established that LLMs struggle to express their intrinsic uncertainty by using epistemic markers in a human-aligned fashion. For example, \citet{metafaith} present the first large-scale systematic study of this phenomenon, finding that faithful linguistic confidence expression is difficult and often harmed by traditional factuality-aligned calibration approaches, but is able to be improved in a model and task-agnostic fashion through metacognitive prompting. Likewise, earlier works such as \citet{yona, gm} present small-scale studies of faithful calibration, while \citet{sft} and \citet{ji} devise fine-tuning and steering approaches to improve it, respectively. While such works recognize the misalignment between internal and linguistically expressed confidence of LLMs, their evaluation hinges on human perceptions of linguistic decisiveness, requiring models to align their use of epistemic markers with human understanding in order to achieve truly faithful confidence expression. Importantly, they fail to consider whether this alignment is present if expressed confidence is defined according to a model’s own perception or use. To the best of our knowledge, no work to date has evaluated this possibility. We therefore present in this work the first study of model-centric faithful calibration.

\section{Experimental Details} \label{app:exps}

\subsection{Technical Details}
In our experiments, Gemini models were accessed through the Gemini Developer API, while GPT models were accessed through the OpenAI API. Open-source model experiments were run on a local server with 8xA6000 (48GB) GPUs.

\subsection{Models} \label{app:models}
We evaluate a total of 13 frontier open-source and proprietary models, varying in size, family, and post-training: Gemini-3.1-Pro \cite{gemini31pro_modelcard_2026}, Gemini-3-Flash \cite{gemini3flashcard}, Gemini-2.5-Flash \citep{gemini25flashcard}, GPT-5(-Mini/Nano) \cite{gpt5card}, Qwen3 (0.6B, 1.7B, 4B, 8B, 32B) \cite{qwen3card}, Llama3.1-Instruct (8B) \citep{grattafiori2024llama3herdmodels}, and Llama3.3-Instruct (70B). 

Similar to prior work \cite{metafaith}, for all models we set the maximum output length to 256 tokens to balance answer completeness and succinctness and use a temperature of 1.0 unless otherwise specified. We do not use thinking mode for any model as our focus is on the reliability of models’ use of epistemic markers when conveying their answer to a query, and not on task or reasoning performance. However, if reasoning cannot be disabled (e.g., Gemini-3.1-Pro), we use the minimum admissible thinking level and set the maximum output length to 512 tokens to ensure responses are not prematurely truncated. For the Qwen3 family, we use the developer-recommended inference hyperparameters. For Qwen3-4B, we specifically use the \texttt{Qwen/Qwen3-4B-Instruct-2507} version, as it is designated as an updated, stronger edition of the model.

\subsection{Datasets} \label{app:datasets}
To analyze the consistency and reliability of models’ use of epistemic markers to signal intrinsic confidence, we select a suite of 11 datasets, listed below, spanning diverse content domains and task formats, including knowledge-intensive QA, natural language inference, scientific reasoning, commonsense reasoning, and hallucination detection.
While these tasks represent a wide range of difficulty levels, since consistent and reliable uncertainty expression is precisely important in difficult task settings \cite{10.1145/3630106.3658941}, our focus leans toward more challenging datasets for which responses are expected to include epistemic markers.
We limit the train (or test) split of each dataset to at most 5000 randomly sampled examples
since prior work \cite{yona} and our preliminary experiments experiments showed this is sufficient for stable results. Details regarding each dataset are provided below:
\begin{itemize}
\item PopQA \cite{popqa} is a knowledge-intensive QA dataset featuring 14,000 entity-centric QA pairs. It is likely to require LLMs to express uncertainty as it includes many tail entities which are difficult for models to capture. Following \citet{yona, metafaith}, we preprocess the data to keep only the ‘director’, ‘screenwriter’, ‘producer’, ‘author’, ‘place of birth’, and ‘occupation’ relations and remove entities less than two characters in length.
\item SelfAware \cite{selfaware} is a knowledge-intensive QA task consisting of 2337 answerable and 1032 unanswerable questions posed by human users.
\item SimpleQA \cite{simpleqa} is a factuality benchmark curated adversarially against GPT-4 responses to ensure a high level of difficulty. It aims to measure LLMs’ ability to answer short, challenging questions.
\item HaluEval \cite{halueval} is a hallucination evaluation benchmark covering QA, summarization, and knowledge-grounded dialogue tasks.
\item MMLU \cite{mmlu} is a benchmark designed to assess the knowledge and problem-solving abilities of LLMs across 57 tasks and a wide range of content domains.
\item SciQ \cite{sciq} is a dataset of 13,679 crowdsourced science exam questions spanning physics, biology, chemistry, and other subfields, consisting of multiple-choice questions with 4 answer options each.
\item ARC-Challenge is the Challenge Set of the AI2 Reasoning Challenge \cite{arcc}, which consists of 2,590 knowledge-intensive science questions. Versus simple QA, these questions are far more challenging as they require integration of multiple information sources.
\item SuperGLUE \cite{superglue} is a natural language understanding benchmark designed to be more rigorous and challenging than GLUE \citep{wang-etal-2018-glue}.\footnote{We sample equally from the ‘boolq’, ‘copa’, ‘wic’, and ‘wsc’ subsets in our experiments.}
\item AmbigNQ \cite{min-etal-2020-ambigqa} is a dataset of 14,042 annotations on ambiguous questions from NQ-OPEN \cite{kwiatkowski-etal-2019-natural}. We use the disambiguated questions from the \texttt{light} split of the dataset.
\item TruthfulQA \cite{lin-etal-2022-truthfulqa} is a dataset specifically designed to evaluate the truthfulness of LMs in generating answers to a wide range of questions. We use the \texttt{generation} split in our experiments.
\item WNLI \cite{wnli} is a dataset which presents sentence pairs for natural language inference, wherein models are tasked with determining wither the second sentence entail a correct interpretation of the first one.
\end{itemize}

\subsection{Prompts}\label{app:prompts}
To elicit model responses in our main experiments, we use the system prompt shown in Fig. \ref{fig:sys2} and the task-specific user prompts shown in Fig. \ref{fig:taskprompts}. The task prompts employ a shared base query format, differentiated for different task types via addition (or non-use) of a brief description of the expected output. For multiple-choice tasks, we observed in preliminary experiments that permuting answer choices did not affect \micx analysis results. Therefore, we use a randomized answer choice ordering for all multiple-choice datasets. As discussed in \S\ref{subsec:sysprompt}, we also explore the impact of using a metacognitive system prompt to elicit uncertainty expressions that more faithfully reflect models' intrinsic uncertainty; this prompt is shown in Fig. \ref{fig:sys3}. Concretely, the metacognitive prompt additionally instructs the model that it has good \textit{metacognitive sensitivity} and privileged access to its internal uncertainty signals. To score accuracy of model responses, we use the LLM-as-a-Judge with the exact model (Gemini-2.0-Flash\footnote{We verify the efficacy of this LLM judge for accuracy using the same setup as the consistency judgment verification procedure described in \S\ref{app:conf}: using the same 300 randomly selected samples to compare accuracy judgments assigned by Gemini-2.0-Flash versus human annotators, we observed that Gemini-2.0-Flash achieved an accuracy of 0.99, demonstrating the model to be strong and sufficient at this task. Inter-annotator agreement was measured via Krippendorff's alpha to be 1.00.}) and prompt setup from prior work \cite{metafaith}; the prompt is reproduced in Fig. \ref{fig:acc}. To extract epistemic markers from model response sentences and to standardize the extracted epistemic markers by merging those which are essentially the same in form (\S\ref{subsec:extraction}), we use Gemini-2.5-Flash-Lite with the prompts shown in Fig.s \ref{fig:hedgeextractionprompt} and \ref{fig:hedgestandardizationprompt} with greedy decoding and designated stop sequences ``Answer:’’ and ``\}'', respectively.
\begin{figure}[t]
\centering
\begin{tcolorbox}[colframe=black, colback=gray!5, boxrule=0.5pt, arc=2mm, width=\textwidth, left=1mm, right=1mm, top=1mm, bottom=1mm,title=Standardized System Prompt]
Your task is to provide a succinct and accurate answer to the given question. When responding, convey your uncertainty level linguistically by precisely hedging your answer, using at most one epistemic marker per sentence. Provide only your hedged answer without asking anything of the user. Do not ask anything of the user. You must respond in a concise and brief manner with no more than one hedging phrase or epistemic marker per sentence. Limit your response to a single sentence or phrase if possible, or use at most 2-3 sentences.
\end{tcolorbox}
\caption{System prompt used to elicit uncertainty-bearing model responses across all experiments aside from those discussed in \S\ref{subsec:sysprompt}.} \label{fig:sys2}
\vspace{-3mm}
\end{figure}
\begin{figure}[h]
\centering
\begin{tcolorbox}[colframe=black, colback=gray!5, boxrule=0.5pt, arc=2mm, width=\textwidth, left=1mm, right=1mm, top=1mm, bottom=1mm,title=Template for Simple QA Tasks]
\blue{\{question\}}\\
Answer:
\end{tcolorbox}
\begin{tcolorbox}[colframe=black, colback=gray!5, boxrule=0.5pt, arc=2mm, width=\textwidth, left=1mm, right=1mm, top=1mm, bottom=1mm,title=Template for QA Tasks with Unanswerability]
\blue{\{question\}} State that the question is unanswerable if you think it is unanswerable.\\
Answer:
\end{tcolorbox}
\begin{tcolorbox}[colframe=black, colback=gray!5, boxrule=0.5pt, arc=2mm, width=\textwidth, left=1mm, right=1mm, top=1mm, bottom=1mm,title=Template for QA Tasks with Context]
\blue{\{context\}}\\
\blue{\{question\}}\\
Answer:
\end{tcolorbox}
\begin{tcolorbox}[colframe=black, colback=gray!5, boxrule=0.5pt, arc=2mm, width=\textwidth, left=1mm, right=1mm, top=1mm, bottom=1mm,title=Template for Multiple-Choice Tasks]
\blue{\{question\}}\\
Choices: \blue{\{choices\_list\}}\\
Answer:
\end{tcolorbox}
\begin{tcolorbox}[colframe=black, colback=gray!5, boxrule=0.5pt, arc=2mm, width=\textwidth, left=1mm, right=1mm, top=1mm, bottom=1mm,title=Template for NLI Tasks]
Premise: \blue{\{text\_1\}}\\
Hypothesis: \blue{\{text\_2\}}\\
Does the premise entail the hypothesis? (yes or no)\\
Answer:
\end{tcolorbox}
\begin{tcolorbox}[colframe=black, colback=gray!5, boxrule=0.5pt, arc=2mm, width=\textwidth, left=1mm, right=1mm, top=1mm, bottom=1mm,title=Template for Hallucination Detection Tasks]
Question: \blue{\{question\}}\\
Response: \blue{\{response\}}\\
Does the response contain hallucination? (yes or no)\\
Answer:
\end{tcolorbox}
\caption{Task-specific prompts used to elicit model responses across experimental settings.} \label{fig:taskprompts}
\end{figure}

\begin{figure}[h]
\centering
\begin{tcolorbox}[colframe=black, colback=gray!5, boxrule=0.5pt, arc=2mm, width=\textwidth, left=1mm, right=1mm, top=1mm, bottom=1mm,title=Metacognitive System Prompt (\S\ref{subsec:sysprompt})]
You are an agent with high metacognitive sensitivity and excellent self-awareness of your internal confidence and uncertainty. Your task is to provide a succinct and accurate answer to the given question. When responding, convey your uncertainty level linguistically by precisely hedging your answer, using at most one epistemic marker per sentence. Provide only your hedged answer without asking anything of the user. Do not ask anything of the user. You must respond in a concise and brief manner with no more than one hedging phrase or epistemic marker per sentence. Limit your response to a single sentence or phrase if possible, or use at most 2-3 sentences.
\end{tcolorbox}
\caption{Metacognitive system prompt adapted from \citet{metafaith}, used to elicit model responses which express internal uncertainty in a more human-aligned fashion.} \label{fig:sys3}
\end{figure}

\begin{figure}[h]
\centering
\begin{tcolorbox}[colframe=black, colback=gray!5, boxrule=0.5pt, arc=2mm, width=\textwidth, left=1mm, right=1mm, top=1mm, bottom=1mm,title=Accuracy Scoring Prompt]
Determine whether the predicted answer contains text semantically equivalent to any of the ground truth options. Output ONLY True or False.\\
ground truth options = \textcolor{blue}{\{targets\}}\\
predicted answer = \textcolor{blue}{\{pred\}}
\end{tcolorbox}
\caption{Prompt used to score correctness of model responses via LLM-as-a-Judge.} \label{fig:acc}
\end{figure}

\begin{figure*}[t]
\centering
\footnotesize
\begin{tcolorbox}[colframe=black, colback=gray!5, boxrule=0.5pt, arc=2mm, width=\textwidth, left=1mm, right=1mm, top=1mm, bottom=1mm,title=Prompt to Extract LLM-Emitted Epistemic Markers]
You are a linguistic expert. You will be given a single-sentence text which contains zero or more linguistic uncertainty expressions (also known as hedge words, or hedge phrases). Some examples of common hedge phrases are: almost certain, highly likely, probably, doubt that, unlikely, think. Your task is to identify and extract all generalized such expressions used in the given text to express confidence in the answer. Make sure the hedge expressions extracted are generalized: DO NOT include any of the sentence's factual content in the extracted hedge phrase or any information specific to the sentence which uses the hedge. Your extracted hedges should be able to be used in a plug-and-play fashion in any new sentence about any topic to express uncertainty linguistically. Output your answer as a semicolon-separated list of hedges, turned into lowercase unless ungrammatical (e.g., 'I' should be capitalized), with no other text. If no hedge is used respond nothing. End your entire response with \#\#\#\#\\

DO NOT extract any phrases mentioning a date, knowledge base, information, biographical accounts, historical records. DO NOT extract “yes” or “no” as a hedge phrase. \\
BAD examples representing the types of hedge phrases you should AVOID extracting include: At the time of that publication (too specific, mentions answer), he is known for (too specific), historical documentation confirms (too specific), Arnold is a possibility (mentions answer), not available in my current knowledge base, most biographical accounts indicate (too specific, suggests answer is a person)\\
These are all NOT generalizable, i.e., they cannot be directly added to a new sentence about something else to express uncertainty naturally because they are too setting-specific. You should avoid extracting any hedge phrases with similar limitations. If you do not adhere to this, you will lose your job.\\

Text: I think the Warcraft wiki says 13,000 years, but I could be mistaken.\\
Hedges: I think; I could be mistaken \#\#\#\#\textbackslash n\\
Text: Based on available information, Islamia College of Science and Commerce was probably the college accredited by the UGC in April 2010. I am not too sure.\\
Hedges: based on the available information; probably; I am not too sure \#\#\#\#\textbackslash n\\
Text: All signs point to Linus Pauling.\\
Hedges: all signs point to \#\#\#\#\textbackslash n\\
Text: The Mediterranean Sea's maximum depth measures 5,109 meters.\\
Hedges: \#\#\#\#\textbackslash n\\
Text: To the best of my recollection, it was May 10, 2023, when they made the announcement.\\
Hedges: to the best of my recollection \#\#\#\#\textbackslash n\\
Text: I believe it was 13 May 2004 when he received the appointment.\\
Hedges: I believe \#\#\#\#\textbackslash n\\
Text: That information isn’t available to me, so I can’t respond with certainty.\\
Hedges: I can't respond with certainty \#\#\#\#\textbackslash n\\
Text: Perhaps he was 80 years old. \\
Hedges: perhaps \#\#\#\#\textbackslash n\\
Text: There’s a chance that the private launch happened on May 10, 1996.\\
Hedges: there's a chance \#\#\#\#\textbackslash n\\
Text: I’m almost certain I’m wrong, but maybe *Human Planet* first aired on 25 April 2011 on Discovery en Español.\\
Hedges: I'm almost certain I'm wrong; maybe \#\#\#\#\textbackslash n\\
Text: I’m fairly sure his birthday is today\\
Hedges: I'm fairly sure \#\#\#\#\textbackslash n\\
Text: It's clear that the American Classical Music Hall of Fame inducted four people this year. I know this for a fact.\\
Hedges: it's clear that; I know for a fact \#\#\#\#\textbackslash n\\
Text: My tentative answer is yes\\
Hedges: my tentative answer is \#\#\#\#\textbackslash n\\
Text: She might have been 50 when she passed away.\\
Hedges: might \#\#\#\#\textbackslash n\\
Text: As of 2022, the answer was no.\\
Hedges: \#\#\#\#\textbackslash n\\
Text: Emmett is a possibility\\
Hedges: is a possibility\\
Text: 1964 is what comes to mind\\
Hedges: is what comes to mind \#\#\#\#\textbackslash n\\
Text: August 2011 sounds familiar\\
Hedges: sounds familiar \#\#\#\#\textbackslash n\\
Text: Historical documentation confirms the answer is 10.\\
Hedges: \#\#\#\#\textbackslash n\\
Text: \blue{\{text\}}\\
Hedges: <your comma-separated list here>
\end{tcolorbox}
\caption{Prompt used to extract hedge expressions and epistemic markers from model response sentences.} \label{fig:hedgeextractionprompt}
\vspace{-3mm}
\end{figure*}

\begin{figure*}[t]
\centering
\footnotesize
\begin{tcolorbox}[colframe=black, colback=gray!5, boxrule=0.5pt, arc=2mm, width=\textwidth, left=1mm, right=1mm, top=1mm, bottom=1mm,title=Prompt to Standardize Extracted Epistemic Markers]
You are a linguistics expert. Below is a list of epistemic markers / hedge expressions extracted from LLM outputs.\\

Your task: strip non-hedging adverbs (e.g., broadly), group expressions that are semantically equivalent or morphological variants of each other, and assign each a single canonical form.\\

Rules:\\
- Canonical form should be the simplest/shortest representative (e.g. "not certain" not "I am not certain") with minimal adjectives and adverbs\\
- Consider "not \_\_\_" and "un\_\_\_" instances as distinct, e.g. "not certain" and "uncertain" should be distinct and not grouped together\\
- Merge morphological variants (`suggests', `suggesting' $\rightarrow$ `suggest')\\
- MERGE expressions with same meaning despite different surface form. Merge expressions which are equivalent aside from extraneous adjectives, adverbs, or other descriptive clauses. Examples:\\
\-\hspace{0.25in}- MERGE "possible" and "possibly"\\
\-\hspace{0.25in}- MERGE "most possibly" and "most likely"\\
\-\hspace{0.25in}    - MERGE "could possibly", "could potentially", "could potentially be", "could potentially have"\\
\-\hspace{0.25in}    - Map "might seem to likely be possibly related" to "might seem"\\
\-\hspace{0.25in}    - Map `may not have up-to-date or comprehensive information', `may not have up-to-date information',`may not have information', `may not have accurate information' all to "may not have information"\\
\-\hspace{0.25in}    - Map "couldn't verify for certain", "couldn't verify with certainty", "couldn't verify with absolute certainty", "couldn't verify accuracy", "couldn't verify certainty" all to "couldn't verify"\\
- Do NOT merge expressions with meaningfully different confidence levels (e.g. "probably" vs "possibly")\\
- If an expression is already canonical, map it to itself\\
- If anything in the input is not a hedge, e.g. empty string "" or meaningless text such as "<answer>", map it to itself\\

Return ONLY a valid JSON object mapping EACH input expression to its canonical form, like:\\
\{\{"I am not certain": "not certain", "suggests": "suggest", "suggesting that": "suggest", "I am not aware of any alternative": "not aware", "I am not aware": "not aware", "I am not aware of": "not aware", "I am fairly certain": "fairly certain", "I am fairly certain about this information": "fairly certain", "I am fairly confident": "fairly confident", 'appears to have been': `appears to have been', `appears to have been likely': `appears to have been', `appears to have been written by': `appears to have been', `appears to have likely been': `appears to have been', `appears to have possibly': `appears to', `appears to likely': `appears to', `appears to likely be': `appears to', `appears to possibly be': `appears to', `appears to probably': `appears to', `may not have up-to-date or comprehensive information': "may not have information", `may not have up-to-date information': "may not have information", `may not have information': "may not have information", `may not have accurate information': "may not have information", "couldn't verify for certain": "couldn't verify", "couldn't verify with certainty": "couldn't verify", "couldn't verify with absolute certainty": "couldn't verify", "couldn't verify accuracy": "couldn't verify", "couldn't verify certainty": "couldn't verify", `may be argued': `may be', `may be assumed': `may be', `may be considered': `may be', `may be incomplete': `may be', `may be likely': `may be', `may be limited': `may be', `may be limited or outdated': `may be', ...\}\}\\
Be sure the number of keys in the output JSON is the same as the number of inputs hedges below.\\

Hedge expressions:\\
\blue{\{extracted\_markers\_list}\}
\end{tcolorbox}
\caption{Prompt used to standardize the format of epistemic markers extracted from model response sentences.} \label{fig:hedgestandardizationprompt}
\vspace{-3mm}
\end{figure*}

\subsection{Quantifying Intrinsic Confidence} \label{app:conf}
\begin{figure}[t]
\centering
\begin{tcolorbox}[colframe=black, colback=gray!5, boxrule=0.5pt, arc=2mm, width=\textwidth, left=1mm, right=1mm, top=1mm, bottom=1mm,title=Consistency Judgment Prompt]
Context: \textcolor{blue}{\{sampled\_response\}}\\
Assertion: \textcolor{blue}{\{sentence\}}\\
Is the assertion consistent with the context above?\\
Answer Yes or No:
\end{tcolorbox}
\caption{Prompt \cite{metafaith, manakul-etal-2023-selfcheckgpt} used to assess sentence-response consistency when estimating models' intrinsic confidence.}\label{fig:confprompt}
\vspace{-3mm}
\end{figure}
As discussed in \S\ref{sec:mic}, we follow previous work to quantify models' intrinsic confidence by assessing consistency across sampled responses. 
To do this, we use the exact methodology of \citet{metafaith}, which is adapted from \citet{manakul-etal-2023-selfcheckgpt} and, unlike the method proposed by \citet{yona}, does not depend on having the same number or order of assertions among sampled responses. Given a text input $Q$ and response $R=\{s_1,\ldots, s_L\}$, an additional $K=20$\footnote{We use $K=20$ as existing work \citep{manakul-etal-2023-selfcheckgpt, tian2024finetuning} shows going beyond this number yields marginal returns on estimate quality. In general, however, $K=10$ sufficient for similar paradigms \citep{kuhn2023semantic, chen-mueller-2024-quantifying, rivera-etal-2024-combining}.} responses $R_1,\ldots,R_K$ are sampled. The consistency between each sentence $s_l$ and response $R_k$ is then assessed by querying Gemini-2.0-Flash\footnote{We use Gemini-2.0-Flash as it is deemed sufficiently capable in prior work given the simplicity of the task, since it has superior capabilities to GPT-3, which was found to be an effective judge LLM by \citet{manakul-etal-2023-selfcheckgpt}, and in light of our own manual verification, described later in this subsection.} to perform a simple NLI judgment with the prompt shown in Fig. \ref{fig:confprompt}. Judgments are converted to inconsistency scores $x_l^k$ through the mapping \{\text{yes}: 0.0\text{, n/a}: 0.5\text{, no}: 1.0\}, and the overall intrinsic confidence of $M$ in $s_l$ is computed as
\[\texttt{conf}_M(s_l) := 1 - \frac{1}{K} \sum_{k} x_l^k.  \]

To verify the efficacy of this paradigm, we conduct a small-scale human annotation study similar to the one run by \citet{metafaith}. Specifically, we took 300 randomly selected examples by selecting 100 each from the PopQA, SciQ, and AmbigNQ datasets, evenly split across 4 representative models Llama3.1-8B-Instruct, Qwen3-32B, Gemini-2.5-Flash, and GPT-5. This yielded a total of 25 outputs per model per dataset $\times$ 4 models $\times$ 3 datasets = 300 examples for annotation. We then compared consistency judgments from Gemini-2.0-Flash versus annotator-assigned labels for these examples. Annotation was performed by one author and one researcher working in NLP and consisted of determining consistency between sampled responses using the same instruction given to the LLM judge. Given the simplicity and small-scale nature of the task, no special interface was used and no compensation was given; informed consent was obtained prior to collecting annotation results. Inter-annotator agreement was measured via Krippendorff's alpha to be 0.98. Overall, Gemini-2.0-Flash achieved an accuracy of 0.98 versus authors, and a high Spearman correlation of 0.99 was observed between the confidence scores resulting from each approach. Given the near-perfect performance of Gemini-2.0-Flash for this task, we deemed it to be sufficient.

\subsection{Quantifying Linguistic Decisiveness} \label{app:dec}
As part of our analysis (\S\ref{sec:metrics}, \S\ref{sec:results}) depends on sentence-level comparison of models' intrinsic confidence against the human-perceived linguistic decisiveness of their generations, we follow the precedent of prior work \cite{yona, ji, metafaith, sft} to evaluate decisiveness in a human-aligned fashion via LLM-as-a-Judge. In particular, we use the prompt from \citet{metafaith} shown in Fig. \ref{fig:decprompt} to instruct Gemini-2.0-Flash to assign decisiveness scores between 0 and 1 for each sentence. 

We reproduce the verification experiments by \citet{metafaith} to confirm the resulting judgments are well-aligned with human annotations.
In particular, manual verification by authors for decisiveness is unreliable as it would simply provide single-point estimates of human-interpreted decisiveness which is known to be variable among individuals. Thus, we systematically compared LLM-assigned decisiveness scores against aggregated human annotations in two ways.

First, we utilize a dataset of 800 texts spanning various lengths and multiple domains collected by \citet{gm}. Each text is paired with aggregated human-rated decisiveness scores, which were used to compute the Pearson correlation, Spearman correlation, and standardized mean-squared error (MSE) between LLM ratings and average human ratings reported by \citet{gm}. We compared multiple judge models for this task and report in Table \ref{tab:dec_judges} the results of the best inference temperature each out of 0.0, 0.5, and 1.0; all correlations are significant with $p<0.05$. It can be seen that Gemini-2.0-Flash achieves the highest correlations, confirming the quality of the decisiveness scores used in our work. Additionally, it achieves the lowest MSE of 0.635, better than the MSE of approximately 0.72 observed by \citet{gm} when they use GPT-4o fine-tuned on human ratings to obtain decisiveness scores. Other judgment prompts were explored per model but did not substantially alter the findings.
\begin{table}[t]
\centering\small
\caption{Relative efficacy of different LLM judges for linguistic decisiveness.}
\begin{tabular}{lccc}\toprule
Model & Pearson Corr. & Spearman Corr. & MSE \\ \midrule
GPT-5 & 0.554 & 0.534 & 0.89 \\
GPT-4o & 0.639 & 0.620 & 1.78 \\
GPT-4o-Mini & 0.514 & 0.544 & 0.97 \\
\textbf{Gemini-2.0-Flash} & \textbf{0.680} & \textbf{0.663} & \textbf{0.64} \\
Gemini-2.5-Flash & 0.617 & 0.578 & 0.75 \\
Gemini-2.5-Flash-Lite & 0.599 & 0.544 & 0.77 \\
Gemini-3-Flash & 0.617 & 0.475 & 1.04 \\
Gemini-3.1-Flash-Lite & 0.649 & 0.528 & 0.94 \\ \bottomrule
\end{tabular}
\label{tab:dec_judges}
\end{table}

\begin{table*}[t]
\centering\small 
\caption{Comparison of our mean decisiveness scores for common hedge words vs. the median and IQR of human perceptions of probability \citep{FU}.}
\begin{tabular}{lcc}
\toprule
Hedge Word & Human-Annotated Median (IQR) & Our Mean Decisiveness \\
\midrule
``Almost No Chance'' & 0.02 (0.01, 0.05) & 0.03 \\
``Highly Unlikely''  & 0.05 (0.05, 0.10) & 0.06 \\
``Improbable''       & 0.10 (0.05, 0.22) & 0.12 \\
``Little Chance''    & 0.10 (0.05, 0.15) & 0.14 \\
``Chances are Slight'' & 0.10 (0.10, 0.20) & 0.15 \\
``Unlikely''         & 0.20 (0.10, 0.30) & 0.20 \\
``We Doubt''         & 0.20 (0.10, 0.30) & 0.23 \\
``Probably Not''     & 0.25 (0.15, 0.30) & 0.33 \\
``About Even''       & 0.50 (0.50, 0.50) & 0.55 \\
``Better than Even'' & 0.60 (0.55, 0.60) & 0.64 \\
``Likely''           & 0.70 (0.65, 0.75) & 0.71 \\
``Probably''         & 0.70 (0.60, 0.75) & 0.68 \\
``We Believe''       & 0.75 (0.65, 0.85) & 0.75 \\
``Very Good Chance'' & 0.80 (0.75, 0.90) & 0.75 \\
``Highly Likely''    & 0.90 (0.80, 0.95) & 0.90 \\
``Almost Certain''   & 0.95 (0.90, 0.98) & 0.93 \\
\bottomrule
\end{tabular}
\label{tab:FU_verif}
\end{table*}
We further confirm the efficacy of the final decisiveness scoring setup by randomly sampling 110 sentences each containing one of the 16 human-annotated hedge expressions from \citet{FU}, split evenly across datasets (11 datasets $\times$ 10 samples), for a total of 110 $\times$ 16 = 1760 sentences. We then score each sentence’s decisiveness using our setup and compute the mean decisiveness score issued for sentences with each hedge; these scores are compared against the distribution of human-perceived probabilities \citep{FU} for each hedge. Results are reported in the Table \ref{tab:FU_verif}. It can be seen that our scores track the human-annotated judgments with high consistency, capturing decisiveness in a human-aligned fashion across the whole range.

\begin{figure*}[t]
\footnotesize
\begin{tcolorbox}[colframe=black, colback=gray!5, boxrule=0.5pt, arc=2mm, width=\textwidth, left=1mm, right=1mm, top=1mm, bottom=1mm,title=Decisiveness Scoring Prompt]
We will present you with a statement. Your task is to evaluate the linguistic assertiveness of it. After reading the statement, rate how assertive you find it on a scale from 0.0 (Not at all assertive) to 1.0 (Extremely assertive). Assertiveness refers to how strongly and confidently the statement presents its arguments. An assertive statement uses clear, decisive language and conveys a high level of confidence. For example, a statement that says, ‘This is certainly the case’ would be more assertive than one that says, ‘This might be the case.’. Please be consistent in your ratings. A ‘0.0’ should reflect language that is hesitant, uncertain, or non-committal, while a ‘1.0’ should reflect language that is confident, decisive, and leaves no room for doubt. You should pay special attention to the usage of any hedging modifiers in the original answer, used to convey uncertainty in the truthfulness of the assertion. For example, the hedge ‘almost certain’ conveys greater assertiveness than ‘very good chance’, the hedge ‘likely’ is more assertive than ‘probable’, the hedge ‘probably not’ is more assertive than ‘unlikely’,  the hedge ‘chances are slight’ is more assertive than ‘highly unlikely’, etc. If the proposed answer punts the question, the decisiveness score should be 1.0. ONLY PROVIDE A FLOAT BETWEEN 0.0 AND 1.0 AS YOUR RESPONSE.\textbackslash n

Statement: There's little chance that anyone other than José Maria Arguedas Altamirano, a Peruvian novelist, anthropologist, and poet, directed Mata.\\
Rating: 0.129\textbackslash n

Statement: We doubt anyone knows for certain who the screenwriter for *Two* was, but it may have been Rob Reiner.\\
Rating: 0.224\textbackslash n

Statement: Joris Delle was probably born in Épinal, France.\\
Rating: 0.689\textbackslash n

Statement: The odds of Carol Reed being the director of The Key are about even.\\
Rating: 0.506\textbackslash n

Statement: There was no convincing evidence that delaying the introduction of these foods would prevent allergies.\\
Rating: 0.504\textbackslash n

Statement: supported\\
Rating: 1.0\textbackslash n

Statement: Irreversible losses of decomposition pathways are likely to play key roles in the evolutionary stability of these ubiquitous mutualisms.\\
Rating: 0.267\textbackslash n

Statement: Are you open to changing your view? What would it take to change your view?\\
Rating: 0.011\textbackslash n

Statement: Simpson lives in a fact-free world. Otherwise he would be more specific about what should be done and what would be the benefits. Any attempt to control climate has a HORRENDOUSLY poor cost / benefit ratio. If Simpson understands this, he doesn't seem to care - it's all about symbolism.\\
Rating: 0.779\textbackslash n

Statement: The answer is 100.\\
Rating: 1.0\textbackslash n

Statement: There is not enough information\\
Rating: 1.0\textbackslash n

Statement: \textcolor{blue}{\{text\}}\\
Rating:
\end{tcolorbox}
\caption{Prompt \cite{metafaith} used to score linguistic decisiveness of model responses in a human-aligned fashion via LLM-as-a-Judge.}\label{fig:decprompt}
\vspace{5mm}
\end{figure*}

\subsection{Quantifying Faithful Calibration} \label{app:cmfg}
Faithful calibration refers to the alignment between a model's linguistically expressed and intrinsic confidence. It is based on the faithfulness of models' communicated uncertainty and therefore differs significantly from traditional notions of calibration, which instead aim to align confidence with externally-judged factuality or performance.
Faithful calibration is typically \cite{yona, metafaith, sft} evaluated at the response level by aggregating over assertion- or sentence-level comparisons of intrinsic confidence and human-perceived linguistic decisiveness. In particular, given a query $Q$ and a response $R=\{s_1,\ldots, s_L\}$ generated by a model $M$, the degree to which $R$ is faithful to $M$’s intrinsic confidence is quantified as:
\begin{align*}
    F^M_{Q, R} := 1 - \frac{1}{L}\sum_{l=1}^{L} |\texttt{dec}(s_l) - \texttt{conf}_M(s_l) |.
\end{align*}
Here, $\texttt{dec}(s_l)\in[0,1]$ represents the decisiveness of sentence $s_l$, and $\texttt{conf}_M(s_l)\in[0,1]$ represents $M$'s intrinsic confidence in $s_l$; these are quantified as discussed in \S\ref{app:dec} and \S\ref{app:conf}, respectively. A baseline faithfulness score of 0.5 is achieved if decisiveness is completely unrelated with confidence, and a maximal faithfulness score of 1 is obtained if there is perfect alignment.

Dataset-level faithful calibration is measured by aggregating $F^M$ scores across samples in dataset $D$ using the \cmfgx metric \cite{yona}:
\begin{equation}
    \cmfg_{M,D} := \underset{\substack{i\in [N]\\ c\sim U[0,1]}}{\mathbb{E}}\left[ F^M_{Q_i, R_i} | \texttt{conf}_M(R_i)=c\right] \label{eq:cmfg}
\end{equation}
Here, $N$ is the number of samples in $D$. By conditioning on intrinsic confidence, the $\cmfg$ score controls for variations in confidence score distribution between models to obtain a more reliable estimate of faithful calibration than simple averaging. Following \citet{yona}, we condition over 10 equally sized bins and exclude from computation any samples for which models punt (i.e., decline to answer) the question.

\subsection{Additional Verification Experiments} \label{app:addl_verif}

\begin{table*}[t]
\centering\small\setlength{\tabcolsep}{2pt}
\caption{Average number of hedges used per sentence with vs.\ without specification of using one marker per sentence for representative models.}
\resizebox{\linewidth}{!}{%
\begin{tabular}{lccccccccccc}
\toprule
Model & PopQA & SelfAware & SimpleQA & AmbigNQ & TruthfulQA & HaluEval & WNLI & MMLU & SciQ & ARC-C & SuperGLUE \\
\midrule
Llama3.1-8B-Ins  & 1.03\tiny{±0.12} & 1.01\tiny{±0.40} & 1.08\tiny{±0.36} & 1.14\tiny{±0.82} & 1.30\tiny{±0.43} & 1.00\tiny{±0.42} & 0.99\tiny{±0.34} & 0.75\tiny{±0.56} & 1.18\tiny{±0.32} & 0.92\tiny{±0.33} & 0.89\tiny{±0.25} \\
Qwen3-32B        & 1.32\tiny{±0.17} & 1.16\tiny{±0.09} & 1.22\tiny{±0.67} & 1.12\tiny{±0.37} & 1.20\tiny{±0.67} & 1.11\tiny{±0.25} & 1.21\tiny{±0.28} & 1.30\tiny{±0.53} & 1.15\tiny{±0.41} & 1.38\tiny{±0.52} & 1.31\tiny{±0.57} \\
GPT-5            & 1.04\tiny{±0.19} & 1.21\tiny{±0.56} & 1.03\tiny{±0.18} & 1.04\tiny{±0.21} & 1.24\tiny{±0.52} & 1.26\tiny{±0.38} & 1.05\tiny{±0.29} & 1.38\tiny{±0.34} & 1.11\tiny{±0.43} & 1.08\tiny{±0.19} & 1.17\tiny{±0.17} \\
\bottomrule
\end{tabular}
}
\label{tab:hedges_per_sentence}
\end{table*}

As described in \S\ref{sec:mic}, we assume when computing \micx values that each sentence generated by a model contains at most one epistemic marker. We demonstrate empirically that this requirement is not overly artificial by comparing the average number of hedges used per sentence with versus without such specification for several models. In particular, Table \ref{tab:hedges_per_sentence} reports the mean and standard deviation of the number of hedges used per sentence with versus without specification of using one marker per sentence for 3 representative models on each task used in our experiments. To do this, we remove from the system prompt (Fig. \ref{fig:sys2}) the phrases ``using at most one epistemic marker per sentence'' and ``with no more than one hedging phrase or epistemic marker per sentence.'' It can be seen that even without our enforcement of this quality, models for the most part tend to use 1 marker per sentence, validating the uncertainty expression setup.

Additionally, in \S\ref{subsec:threshold}, we investigate the impact of the marker frequency threshold $T$ on our primary conclusions from \S\ref{subsec:mainresults} by repeating the main experiments using thresholds of $T=20, 50, 100$. These thresholds were determined in a principled fashion, with $T=100$ representing the approximately 30th percentile of marker frequency across all models, markers, and tasks, $T=10$ the basis for prior work \citep{marconf} investigating how well models associate markers with \textit{accuracy,} and intermediate levels of $T=20$ and $T=50$ approximately doubling their respective preceding frequencies. We confirm that the selected thresholds are appropriate (e.g., to ensure $T=100$ is not too high) by computing the mean and standard deviation of $S$ across markers and datasets for each $T$. Recall that $S$ is the total number of sentences containing a marker for a given model and dataset combination, i.e., the number of sentences averaged over when computing a \micx score (Eq. \ref{eq:mic}). Results are shown in Table \ref{tab:t_verif} for three representative models. It can be seen that the number of sentences $S$ used to derive \micx values is quite high in all settings, showing that higher $T$ values does not overly limit the underlying computation set. 

\begin{table}[t]
\centering\footnotesize\setlength{\tabcolsep}{4pt}
\caption{Mean and standard deviation of $S$ (Eq. \ref{eq:mic}) across markers and datasets for each $T$ for representative models.}
\begin{tabular}{lcccc}
\toprule
Model & $T=10$ & $T=20$ & $T=50$ & $T=100$ \\
\midrule
Llama3.1-8B-Ins & 498\tiny{±390} & 448\tiny{±350} & 602\tiny{±462} & 602\tiny{±462} \\
Qwen3-32B       & 303\tiny{±237} & 374\tiny{±247} & 548\tiny{±199} & 668\tiny{±160} \\
GPT-5           & 524\tiny{±261} & 524\tiny{±261} & 524\tiny{±261} & 524\tiny{±261} \\
\bottomrule
\end{tabular}
\label{tab:t_verif}
\end{table}

\subsection{Evaluation Metrics} \label{app:metrics}
As discussed in \S\ref{sec:metrics}, we propose a suite of seven metrics to evaluate the stability and consistency of each model's marker internal confidences (\mic s) across task settings. This section discusses the implementation and interpretation of these metrics in further detail.\footnote{Datasets with a single split are excluded when computing metrics requiring both train and test sets.}

\paragraph{\imae.} The \emph{in-domain average mean absolute error (MAE)} estimates how well a model's \mic s align with its actual intrinsic confidence on in-distribution test tasks. After a model's \mic s are computed using each dataset's training set $\dtrain$, the \imaex is calculated by generating model responses on the test set $\dtest$ of each dataset, segmenting each test response into sentences, extracting and standardizing epistemic marker(s) per sentence, 
computing for each standardized marker the average across all sentences containing the marker of the absolute error between the model's internal confidence in the sentence and the \micx of the marker, 
averaging across markers, and finally averaging across datasets. Extraction and standardization of markers are implemented as described in \S\ref{subsec:extraction} and \S\ref{app:prompts}, and as before we follow prior work \cite{marconf} to exclude from computation any sentences bearing more than one marker, and designate the use of no marker as its own special marker (\texttt{<no\_hedge>}). The overall \imaex calculation is represented by the following formula:
\begin{align*}
    \imae(M) = \frac{1}{N_d} \sum_D \mae_{M,D}
\end{align*}
where $N_d$ denotes the number of datasets, and 
\begin{align*}
 \mae_{M,D} := \frac{\sum\limits_{E} \frac{\sum_{\substack{s\in R_i,E\in s\\i\in [|D_{\text{test}}|]}} \left|\mic_{E, M, D} - \conf(s))\right|}{S_E}}{|N_{\text{markers}}|}. 
\end{align*}
Here, $E$ denotes an epistemic marker, $N_{\text{markers}}$ denotes the total number of markers, $R_i$ denotes the model's response to sample $x_i\in \dtest$, $s$ denotes a sentence in $R_i$, $\mic_{E,M,D}$ denotes the \micx of marker $E$ for model $M$ calculated over the training set of dataset $D$, and
$S_E:=\sum_{\substack{s\in R_i,i\in [|D_{\text{test}}|]}} \mathds{1}\sbrace{E\in s}$ 
denotes the number of response sentences in which marker $E$ appears.

The \imaex ranges in value from 0 to 1, with lower \imaex scores representing better in-domain train-test transferability of \mic s.
Note that averaging over \textit{markers} for each dataset enables us to bypass potential bias due to different relative marker frequencies across model responses per dataset. However, we also consider two additional variants of \imaex in which the per-dataset $\mae$ component is computed either as the straightforward average \textit{across sentences} of sentence-level absolute errors ($\imae_S$), or as the average \textit{across responses} of per-response average sentence-level absolute errors ($\imae_R$). 
While the $\imae_S$ and $\imae_R$ are susceptible to marker frequency bias, we include them for comparison to \imaex results in \S\ref{app:aggregation}. This allows us to investigate the impact of the granularity at which absolute errors are aggregated.

Following the notational conventions introduced previously, the $\imae_S$ is computed as:
\begin{align*}
    \imae_S(M) = \frac{1}{N_d} \sum_D \mae_{M,D}^S
\end{align*}
where
\begin{align*}
 \mae_{M,D}^S := \frac{\sum\limits_{\substack{s\in R_i\\i\in [|D_{\text{test}}|]}} \left|\mic_{E_{s}, M, D} - \conf(s)\right|}{\sum\limits_{\substack{s\in R_i,i\in [|D_{\text{test}}|]}}\mathds{1}\sbrace{\text{some $E$}\in s}}
\end{align*} and $E_s$ denotes the epistemic marker appearing in sentence $s$.
Likewise, the $\imae_R$ is computed as:
\begin{align*}
    \imae_R(M) = \frac{1}{N_d} \sum_D \mae_{M,D}^R
\end{align*}
where
\begin{align*}
 \mae_{M,D}^R := \frac{\sum\limits_{i\in [|D_{\text{test}}|]} \frac{\sum_{s\in R_i}\left|\mic_{E_{s}, M, D} - \conf(s))\right|}{\sum_{s\in R_i} \mathds{1}\sbrace{\text{some $E$}\in s}}}{|\dtest|}.
\end{align*}

\paragraph{\cmae.} The \emph{cross-domain average MAE} assesses how well a model's \mic s generalize to align with intrinsic confidence on out-of-distribution test tasks. It is computed nearly identically to the \imae, but instead compares \mic s and intrinsic confidences using train and test sets from \textit{different} datasets, and averages over pairs of distinct datasets instead of over singular datasets:
\begin{align*}
    \cmae(M) = \frac{1}{\binom{N_d}{2}} \sum_{D\neq D'} \mae_{M,D,D'}
\end{align*}
where $N_d$ denotes the number of datasets and
\begin{align*}
 \mae_{M,D,D'} := \frac{\sum\limits_{E} \frac{\sum_{\substack{s\in R_i,E\in s\\i\in [|D'_{\text{test}}|]}} \left|\mic_{E, M, D} - \conf(s))\right|}{S'_E}}{|N_{\text{markers}}|}. 
\end{align*} As with the \imaex formulation, here $E$ denotes an epistemic marker, $N_{\text{markers}}$ denotes the total number of markers, $R_i$ denotes the model's response to sample $x_i\in D'_{\text{test}}$, $s$ denotes a sentence in $R_i$, $\mic_{E,M,D}$ denotes the \micx of marker $E$ for model $M$ calculated over the training set of dataset $D$, and 
$S'_E:=\sum_{\substack{s\in R_i,i\in [|D'_{\text{test}}|]}} \mathds{1}\sbrace{E\in s}$ 
denotes the number of test response sentences in which marker $E$ appears.

The \cmaex ranges from 0 to 1, with lower \cmaex indicating better cross-domain transferability of \mic s. As with \imaex, to study the impact of aggregation granularity on results, we additionally consider two variants of \cmaex in which the per-dataset $\mae$ component is instead computed by averaging across sentences ($\cmae_S$) or across responses ($\cmae_R$). These are computed as:
\begin{align*}
    \cmae_S(M) = \frac{1}{\binom{N_d}{2}} \sum_{D\neq D'} \mae_{M,D,D'}^S
\end{align*}
\begin{align*}
 \mae_{M,D,D'}^S := \frac{\sum\limits_{\substack{s\in R_i\\i\in [|D'_{\text{test}}|]}} \left|\mic_{E_{s}, M, D} - \conf(s)\right|}{\sum\limits_{\substack{s\in R_i,i\in [|D'_{\text{test}}|]}}\mathds{1}\sbrace{\text{some $E$}\in s}}
\end{align*}
where $E_s$ denotes the epistemic marker appearing in sentence $s$, and
\begin{align*}
    \cmae_R(M) = \frac{1}{\binom{N_d}{2}} \sum_{D\neq D'} \mae_{M,D,D'}^R
\end{align*}
\begin{align*}
 \mae_{M,D,D'}^R := \frac{\sum\limits_{i\in [|D'_{\text{test}}|]} \frac{\sum_{s\in R_i}\left|\mic_{E_{s}, M, D} - \conf(s))\right|}{\sum_{s\in R_i} \mathds{1}\sbrace{\text{some $E$}\in s}}}{|D'_{\text{test}}|}.
\end{align*}
Comparisons of \cmaex results versus $\cmae_S$ and $\cmae_R$ are provided in \S\ref{app:aggregation}.

\paragraph{\mcv.} The \emph{marker-level average coefficient of variation} assesses the consistency of a model's per-marker \micx values across datasets. It is calculated by computing for a given model $M$ the coefficient of variation (CV) of each shared\footnote{Shared markers are defined as those which appear in a model's responses at least ten times per task. As discussed in \S\ref{subsec:extraction}, this threshold is determined following prior work \cite{marconf}. We investigate the impact of higher thresholds in \S\ref{subsec:threshold}.}  marker's \micx values across datasets, and then averaging the resulting CVs across markers:
\begin{align*}
    \mcv(M) = \frac{\sum\limits_{E} \cv_{E,M}}{N_{\text{shared markers across datasets}}}.
\end{align*}
The CV is computed as
\begin{align*}
    \cv_{E,M} := \frac{\sigma(\{\mic_{E, M, D_i}\}_{i\in[N_d]})}{\mu(\{\mic_{E, M, D_i}\}_{i\in[N_d]})}, 
\end{align*}
where $N_d$ denotes the number of datasets, $E$ denotes a shared marker, and $\sigma(\cdot)$ and $\mu(\cdot)$ respectively denote the standard deviation and mean of the \micx of $E$ with respect to $M$ and each dataset.

While there is no theoretical maximum coefficient of variation, a low \mcvx indicates low dispersion among datasets for each marker's \mic, suggesting a model associates consistent internal confidence to each marker regardless of task setting and reflecting stable cross-dataset \micx values per marker for the model.

\paragraph{\dcv.} The \emph{dataset-level average coefficient of variation} measures the average concentration or dispersion of a model's \micx values \textit{within} a dataset. It is computed by calculating for a given model $M$ the per-dataset coefficient of variation of all markers' \mic s, and then averaging across datasets:
\begin{align*}
    \mcv(M) = \frac{1}{N_d} \sum\limits_{D} \cv_{D,M}.
\end{align*}
The CV is computed as 
\begin{align*}
    \cv_{D,M} := \frac{\sigma(\{\mic_{E_i, M, D}\}_{i\in[N^D_{\text{markers}}]})}{\mu(\{\mic_{E_i, M, D}\}_{i\in[N^D_{\text{markers}}]})},
\end{align*}
where $N^D_{\text{markers}}$ denotes the number of markers appearing in the model's responses for dataset $D$, $E_i$ denotes one such marker, and $\sigma(\cdot)$ and $\mu(\cdot)$ respectively denote the standard deviation and mean of the \micx of all $E_i$'s with respect to $M$ and $D$.

Like \mcv, the \dcvx has no theoretical maximum, but a low \dcvx indicates that \micx values are highly concentrated regardless of task setting, suggesting the model holistically fails to meaningfully differentiate its use of epistemic markers. On the other hand, a high \dcvx suggests the model \textit{is} able to clearly delineate the confidence meanings of different markers in a generalized fashion. 

\paragraph{\mrc.} The \emph{marker rank correlation} measures the consistency of a model's \mic-based marker rankings across datasets. In particular, it evaluates the ranking alignment of markers shared between pairs of datasets and is computed as the Fisher\footnote{We use the Fisher-transformed mean rather than the arithmetic mean when averaging correlation coefficients, as the latter is statistically unsound due to the bounded and nonlinear nature of correlation values \cite{7e2958c8-cf46-3edc-b197-57ee54882a19, Fisher014OT}. While Fisher's transformation is derived for Pearson correlations, it is standard practice to apply it to Spearman correlations as well \cite{silverdunlap, spearman_wiley}.} average Spearman correlation coefficient of \micx values for shared markers across all pairs of datasets:
\begin{align*}
    \mrc(M) = \frac{1}{\binom{N_d}{2}} \sum\limits_{D\neq D'} \rho_{D,D'} 
\end{align*}
where $\rho_{D,D'}$ is the Spearman correlation between $\{\mic_{E_i, M, D}\}_{i\in[N^{D,D'}_s]}$ and $\{\mic_{E_i, M, D'}\}_{i\in[N^{D,D'}_s]}$ for $N^{D,D'}_s$ the number of shared markers in $M$'s responses for datasets $D$ and $D'$ and $E_i$ one such marker.

The \mrcx ranges from $-1$ to 1, with low-magnitude values representing minimal correlation between \mic-defined marker rankings, and high-magnitude values representing nearly identical rankings of markers' associated intrinsic confidence.

\paragraph{\mac.} The \mic-accuracy correlation measures the association between a model's \micx values and accuracy across tasks. 
It is computed as the Fisher average across shared markers of the Pearson correlation coefficient between per-dataset \micx values and per-dataset accuracies:
\begin{align*}
    \mac(M) = \frac{1}{N_{\text{shared markers across datasets}}} \sum\limits_{E} r^{\mac}_{E}. 
\end{align*}
Here, $r^{\mac}_E$ is the Pearson correlation between $\{\mic_{E, M, D_i}\}_{i\in[N_d]}$ and $\{\acc_{M, D_i}\}_{i\in[N_d]}$ for $N_d$ the number of datasets, and $E$ denotes one of $N_{\text{shared markers across datasets}}$ markers shared across the model $M$'s responses for all datasets.

The \macx ranges from $-1$ to 1, with a high \macx indicating the model's \mic s track dataset difficulty to reflect performance (i.e., the model has good internal \textit{sensitivity} of its correctness) without encoding a meaningful or differentiated internal confidence scale, and low (or negative) \macx indicating in contrast that internal confidence levels do not (or inversely) track performance or dataset difficulty. We additionally consider the Spearman-correlation-based analog of \macx, denoted as $\mac_{\text{Spear}}$, and present comparative results to the Pearson formulation in \S\ref{app:corr}.

Interestingly, if high \macx is observed for a model, it suggests that model can achieve good faithful calibration by using linguistic confidence expressions that reflect its task accuracy; this would suggest that factual and faithful calibration can indeed be achieved simultaneously, which is in opposition to prior findings by \citet{metafaith}.

\paragraph{\mcc.} The \mic-\cmfgx correlation measures the cross-dataset association between a model's \micx values and faithful calibration level, serving as the faithfulness-based analog to \macx with \cmfgx (Eq. \ref{eq:cmfg}) substituted for accuracy:
\begin{align*}
    \mcc(M) = \frac{1}{N_{\text{shared markers across datasets}}} \sum\limits_{E} r^{\mcc}_{E}. 
\end{align*}
Here, $r^{\mcc}_E$ is the Pearson correlation between $\{\mic_{E, M, D_i}\}_{i\in[N_d]}$ and $\{\cmfg_{M, D_i}\}_{i\in[N_d]}$ for $N_d$ the number of datasets, and $E$ denotes one of $N_{\text{shared markers across datasets}}$ markers shared across the model $M$'s responses for all datasets.

The \mccx ranges from $-1$ to 1, with high \mccx indicating a model's \mic s are predictive of dataset-level faithful calibration, suggesting internal awareness of linguistic confidence expression ability. On the other hand, a negative \mccx would indicate \mic s are, on average, inversely related to faithful calibration level. As with \macx, we additionally consider the Spearman-correlation-based analog of \mccx, denoted as $\mcc_{\text{Spear}}$, and present comparative results to the Pearson formulation in \S\ref{app:corr}.

\section{Additional Results} \label{app:results}

\subsection{Qualitative Examples}
Understanding the stability and consistency of models' marker use depends on dataset-level analysis, so individual examples may not provide fully representative insights. 
Even so, we provide example-level qualitative results in Tables~\ref{tab:qualitative_qwen} and~\ref{tab:qualitative_gpt5}. For repesentative models Qwen3-32B and GPT-5, we sample three outputs each from a knowledge-intensive QA dataset (PopQA) and a scientific QA dataset (SciQ), and report the model's average internal confidence per output alongside the \micx for each observed marker.

\begin{table*}[t]
\centering\footnotesize\setlength{\tabcolsep}{4pt}
\caption{Qualitative examples for Qwen3-32B.}
\resizebox{\linewidth}{!}{%
\begin{tabular}{clp{7cm}cl p{4cm}}
\toprule
\# & Task & Model Output & Internal Confidence & Observed Markers \& \mic s \\
\midrule
1 & PopQA   & The author of \textit{Pen} is likely unknown or unspecified without additional context. It may be that ``Pen'' refers to a specific book or piece of writing. & 0.76 & likely (0.63), may be (0.51) \\[4pt]
2 & PopQA   & I am uncertain that Veselin Vujović was born in Belgrade, Serbia. & 1.00 & uncertain (0.35) \\[4pt]
3 & PopQA   & To the best of my knowledge, there is no widely recognized film titled \textit{What I Know}. & 0.56 & to my knowledge (0.59) \\[4pt]
4 & SciQ    & It is likely that the gas used in carbonated beverages is carbon dioxide. & 1.00 & likely (0.96) \\[4pt]
5 & SciQ    & The strength of a base when dissolved in water is determined by the amount of hydroxide ions produced, as this directly relates to its ability to neutralize acids. & 0.93 & \texttt{<no\_hedge>} (0.99) \\[4pt]
6 & SciQ    & I am reasonably certain that the diaphragm is the primary muscle involved in inflating and deflating the lungs. & 0.95 & reasonably certain (0.89) \\
\bottomrule
\end{tabular}}
\label{tab:qualitative_qwen}
\end{table*}

\begin{table*}[t]
\centering\footnotesize\setlength{\tabcolsep}{4pt}
\caption{Qualitative examples for GPT-5.}
\resizebox{\linewidth}{!}{%
\begin{tabular}{clp{7cm}cl}
\toprule
\# & Task & Model Output & Internal Confidence & Observed Markers \& \mic s \\
\midrule
1 & PopQA & It was likely Brent. & 1.00 & likely (0.81) \\[4pt]
2 & PopQA & George Martin, as far as I know. & 0.60 & as far as I know (0.75) \\[4pt]
3 & PopQA & I'm not completely sure, but the ``Pilot'' episode of \textit{Lost} was written by J.J.\ Abrams and Damon Lindelof. & 0.80 & not completely sure (0.26) \\[4pt]
4 & SciQ  & Most likely, glucose. & 0.90 & most likely (0.81) \\[4pt]
5 & SciQ  & It begins at the stomach, specifically at the pylorus leading into the duodenum. & 0.80 & \texttt{<no\_hedge>} (0.74) \\[4pt]
6 & SciQ  & Colder oceans are often more hostile to algae and cytoplankton. & 0.25 & often (0.89) \\
\bottomrule
\end{tabular}}
\label{tab:qualitative_gpt5}
\end{table*}

\subsection{\micx Visualizations} \label{app:visualizations}
\begin{figure*}[t]
    \centering
    \begin{subfigure}[t]{\textwidth}
        \centering
        \includegraphics[width=0.95\linewidth]{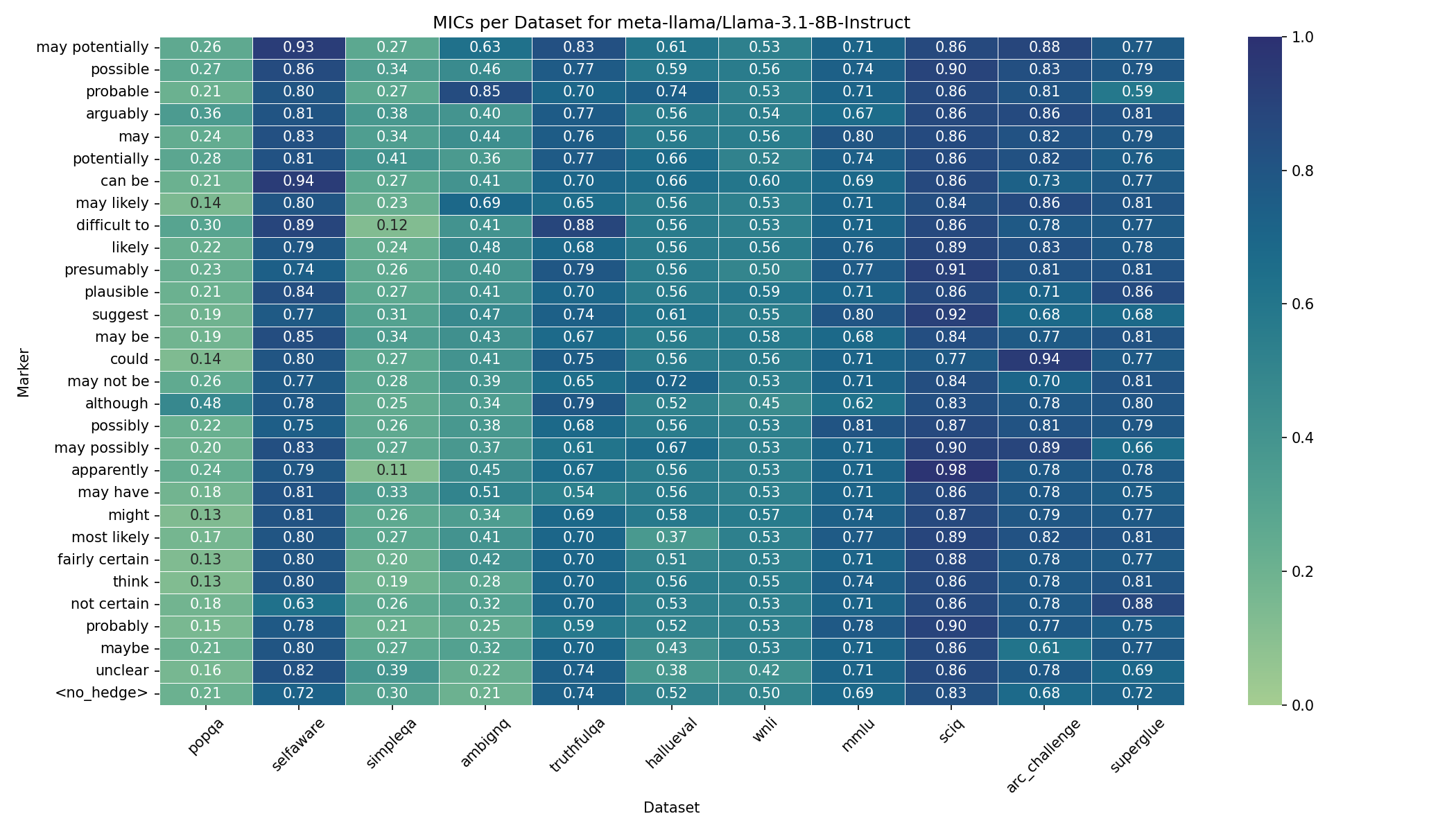}
        \phantomsubcaption\label{fig:llamaheatmap}
    \end{subfigure}%
    \vspace{-6mm}
    \begin{subfigure}[t]{\textwidth}
        \centering
        \includegraphics[width=0.95\linewidth]{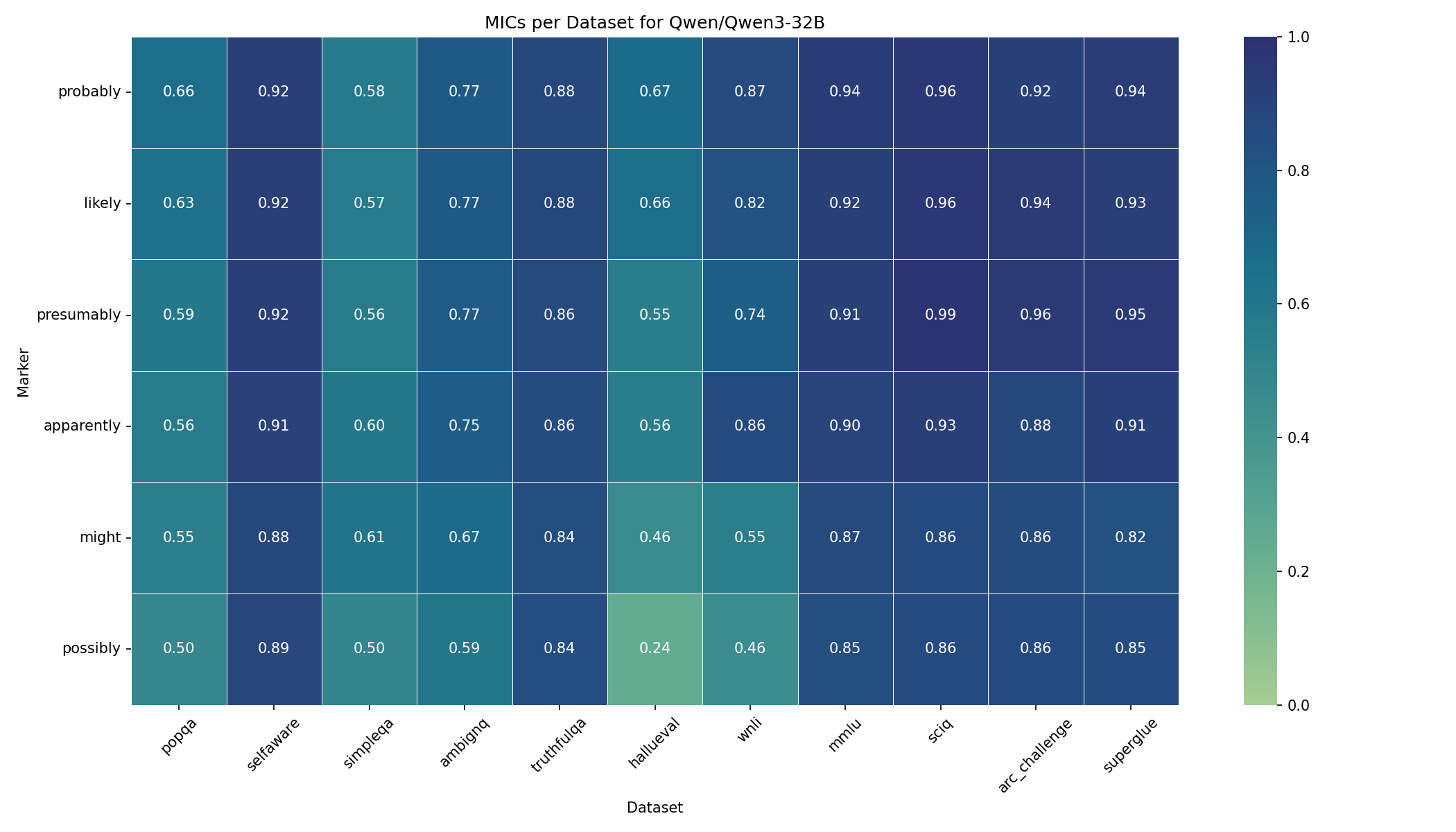}
        \phantomsubcaption\label{fig:qwenheatmap}
    \end{subfigure}
    \vspace{-6mm}\caption{
    \textbf{Representative heatmaps of \micx values for randomly selected epistemic markers across datasets.} 
    We observe that models of varying sizes associate epistemic markers with substantially different internal confidence levels across datasets, often related to task difficulty. Despite this, confidence meanings of individual markers are generally not well-distinguished, consistent with our analysis of the concentration and discriminability of \mic s within and across tasks in \S\ref{sec:results}.
    }\label{fig:heatmaps} 
\end{figure*}

\begin{figure*}[t]
    \centering
    \begin{minipage}[t]{0.75\textwidth}
        \centering
        \includegraphics[width=\linewidth]{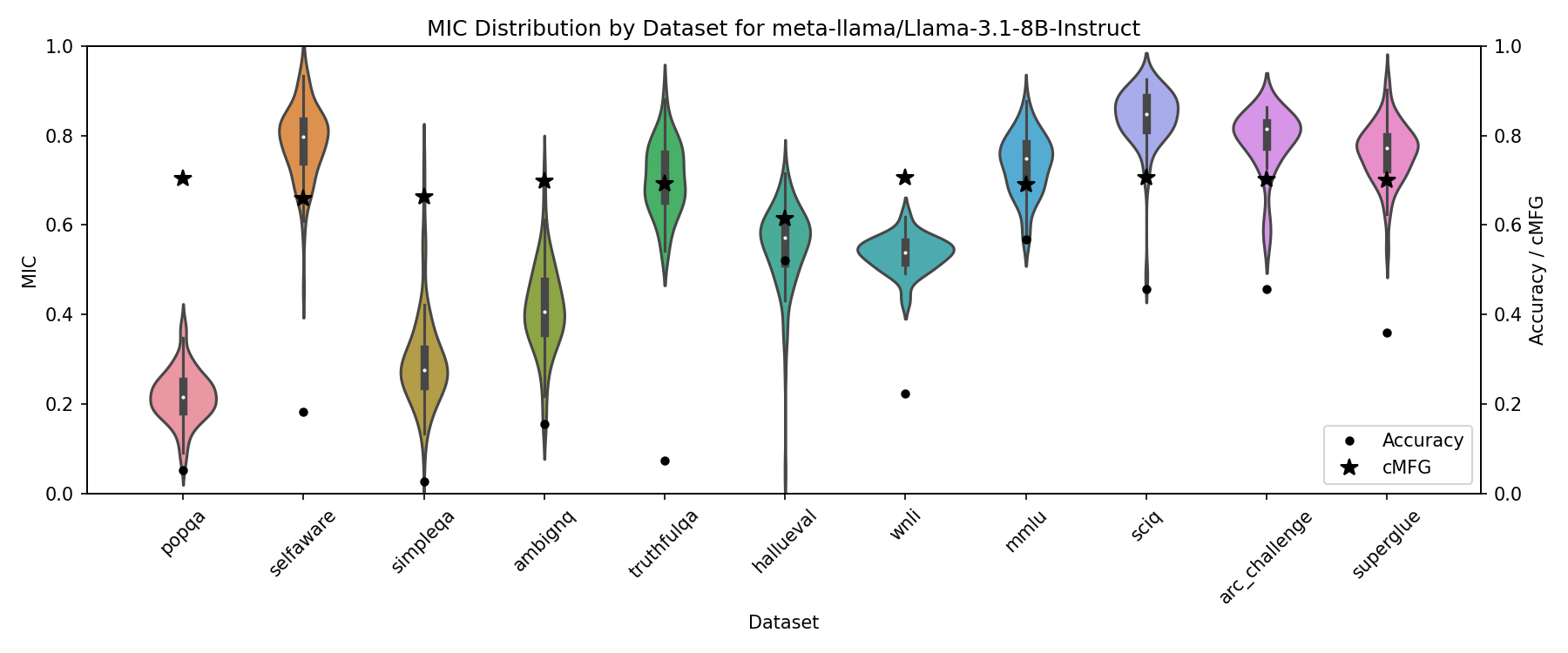}
    \end{minipage}
    \hfill
    \begin{minipage}[t]{0.75\textwidth}
        \centering
        \includegraphics[width=\linewidth]{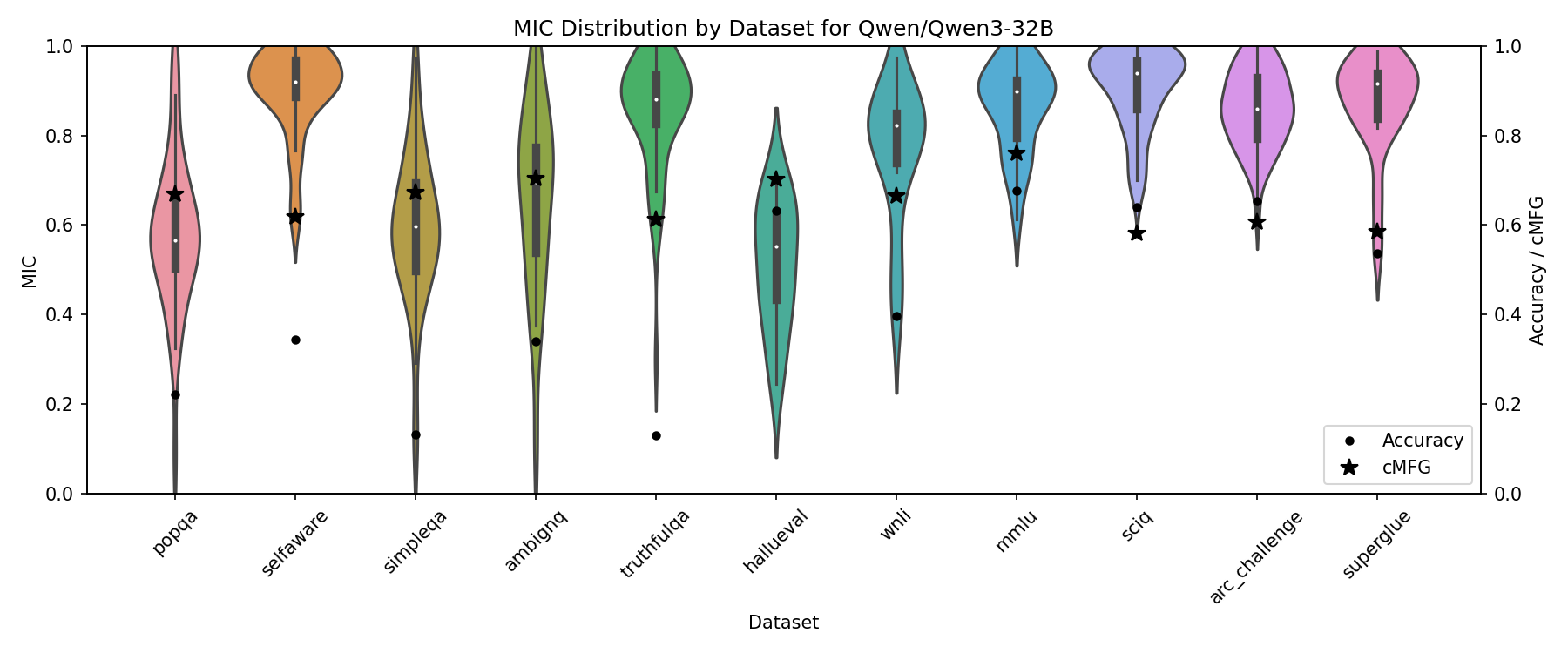}
    \end{minipage}
    \caption{\textbf{Representative violin plots of models' \micx densities across datasets.} Per-dataset accuracy and faithful calibration level (measured via \cmfg) are plotted using dot and star marks, respectively, with values indicated along the second $y$-axis. It can be seen that \micx values sometimes track with accuracy, as suggested by \macx scores in \S\ref{sec:results}.}\label{fig:densities}
\end{figure*}

\begin{figure*}
    \centering
    \includegraphics[width=0.9\linewidth]{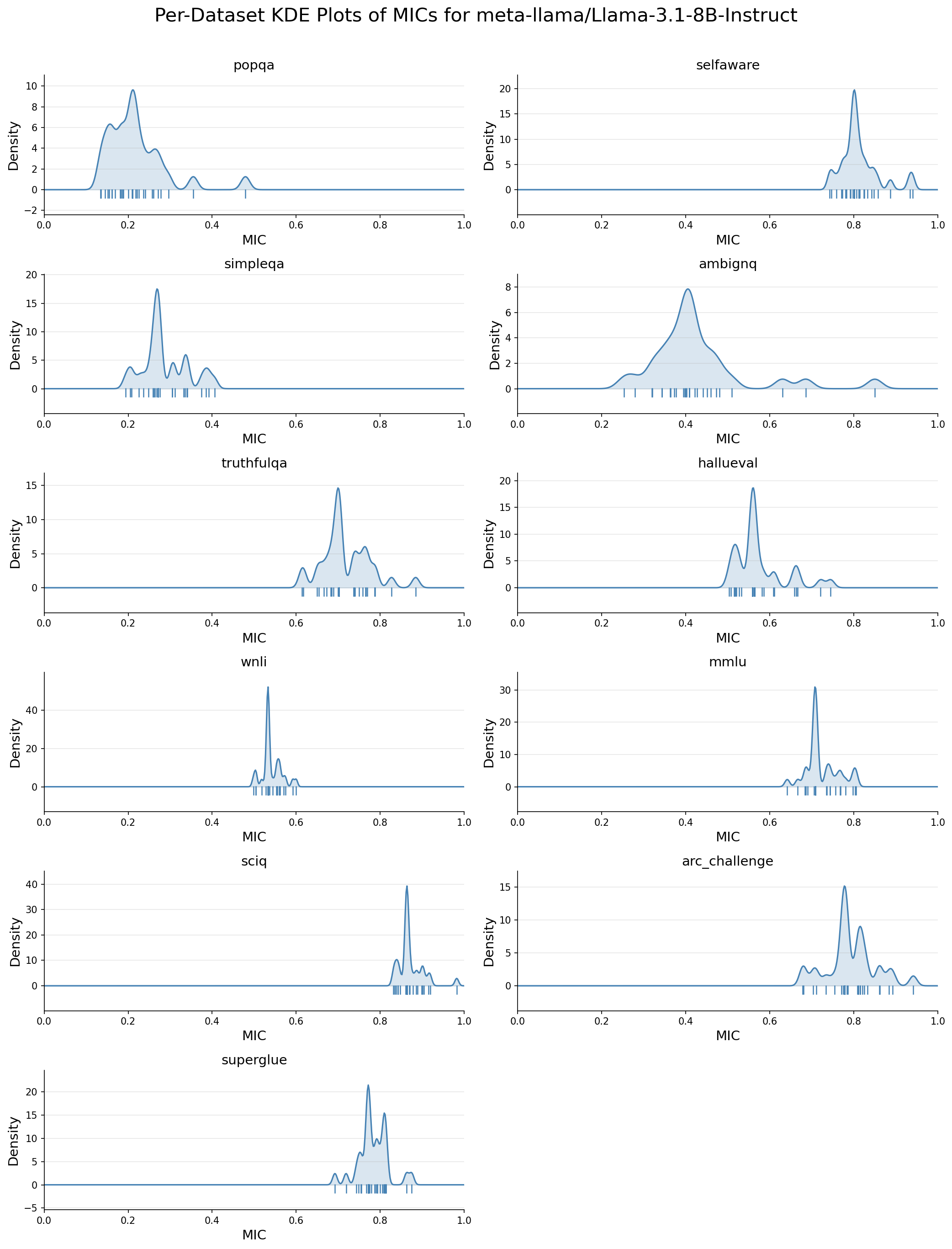}
    \caption{\textbf{Representative KDE plots of \micx values per dataset for Llama3.1-8B-Instruct.} While \micx distributions are generally concentrated within a narrow confidence range, several distinct, albeit weak peaks are observed per task, suggesting the model has some ability to encode multiple distinct uncertainty levels among epistemic markers despite limited marker discriminability.}
    \label{fig:kde1}
\end{figure*}

\begin{figure*}
    \centering
    \includegraphics[width=0.9\linewidth]{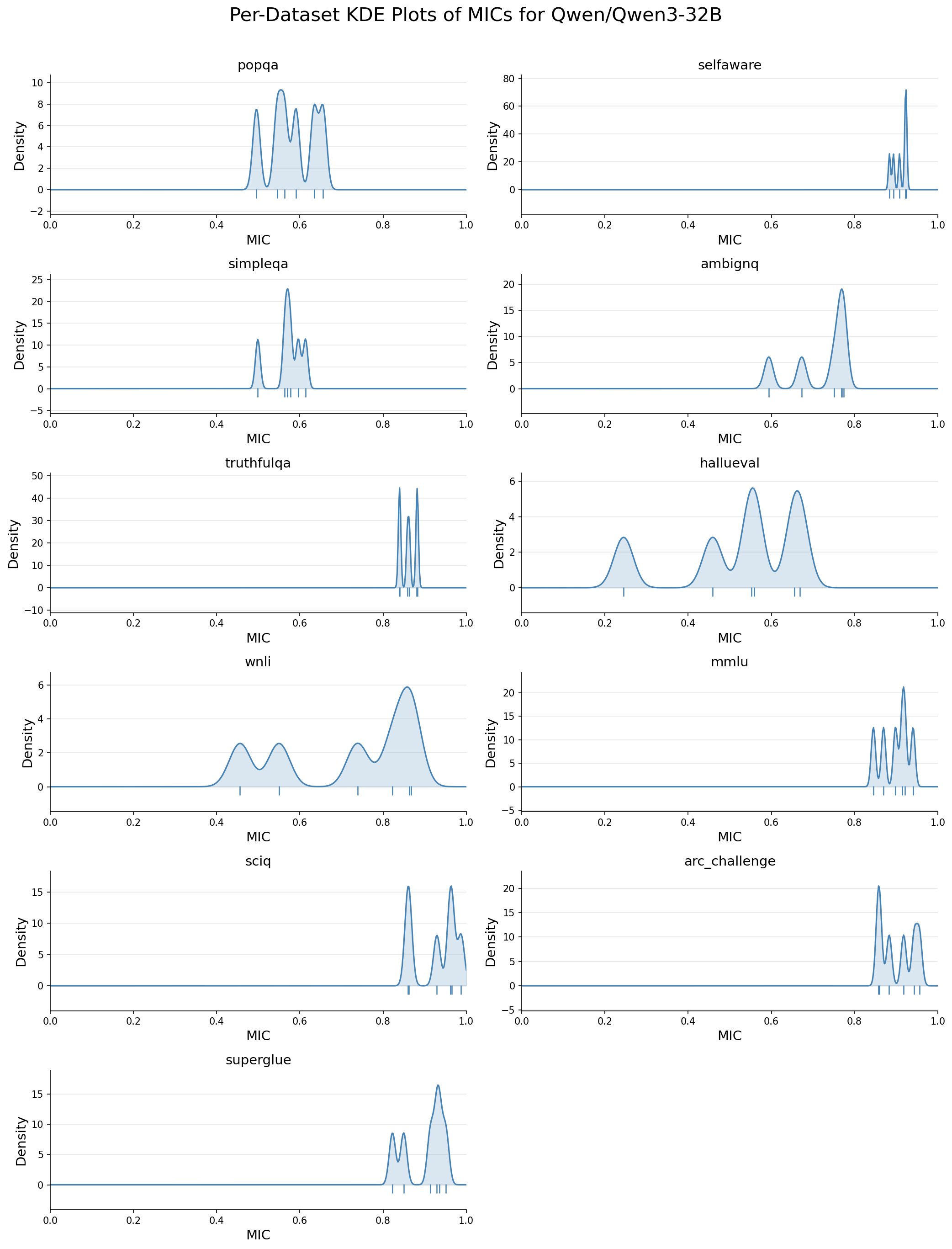}
    \caption{\textbf{Representative KDE plots of \micx values per dataset for Qwen3-32B.} While \micx distributions are generally concentrated within a narrow confidence range, several distinct, often comparable peaks are observed per task, suggesting a multimodal structure with larger models bearing greater ability to encode distinct uncertainty levels among epistemic markers despite limited marker discriminability.}
    \label{fig:kde2}
\end{figure*}

We present representative plots of per-model \micx values and per-model \micx densities across datasets in Fig.s \ref{fig:heatmaps} and \ref{fig:densities}, respectively.
Representative results showing the variance of \micx values are provided in Table \ref{tab:mic_variance}, created by augmenting each \micx value reported for Qwen3-32B in Fig. \ref{fig:heatmaps} with its associated standard deviation.
Representative KDE plots of per-model, per-dataset \mic s are shown in Fig.s \ref{fig:kde1} and \ref{fig:kde2}.

\begin{table*}[t]
\centering\footnotesize\setlength{\tabcolsep}{3pt}
\caption{Representative results showing the variance of \micx values, created by augmenting each \micx value reported for Qwen3-32B in Fig. \ref{fig:heatmaps} with its associated standard deviation. It can be seen that variance is low to moderate, and more challenging tasks are associated with lower \micx values with relatively greater variance.}
\resizebox{\linewidth}{!}{%
\begin{tabular}{lccccccccccc}
\toprule
Hedge & PopQA & SelfAware & SimpleQA & AmbigNQ & TruthfulQA & HaluEval & WNLI & MMLU & SciQ & ARC-C & SuperGLUE \\
\midrule
probably   & 0.66\tiny{±0.18} & 0.92\tiny{±0.06} & 0.58\tiny{±0.16} & 0.77\tiny{±0.12} & 0.88\tiny{±0.09} & 0.67\tiny{±0.24} & 0.87\tiny{±0.15} & 0.94\tiny{±0.14} & 0.96\tiny{±0.10} & 0.92\tiny{±0.97} & 0.94\tiny{±0.03} \\
likely     & 0.63\tiny{±0.17} & 0.92\tiny{±0.06} & 0.57\tiny{±0.16} & 0.77\tiny{±0.12} & 0.88\tiny{±0.09} & 0.66\tiny{±0.17} & 0.82\tiny{±0.11} & 0.92\tiny{±0.18} & 0.96\tiny{±0.10} & 0.94\tiny{±0.03} & 0.93\tiny{±0.03} \\
presumably & 0.59\tiny{±0.16} & 0.92\tiny{±0.06} & 0.56\tiny{±0.16} & 0.77\tiny{±0.11} & 0.86\tiny{±0.12} & 0.55\tiny{±0.19} & 0.74\tiny{±0.10} & 0.91\tiny{±0.19} & 0.99\tiny{±0.04} & 0.96\tiny{±0.09} & 0.95\tiny{±0.01} \\
apparently & 0.56\tiny{±0.13} & 0.91\tiny{±0.07} & 0.60\tiny{±0.13} & 0.75\tiny{±0.18} & 0.86\tiny{±0.10} & 0.56\tiny{±0.21} & 0.86\tiny{±0.14} & 0.90\tiny{±0.20} & 0.93\tiny{±0.08} & 0.88\tiny{±0.08} & 0.91\tiny{±0.06} \\
might      & 0.55\tiny{±0.17} & 0.88\tiny{±0.08} & 0.61\tiny{±0.14} & 0.67\tiny{±0.14} & 0.84\tiny{±0.12} & 0.46\tiny{±0.21} & 0.55\tiny{±0.12} & 0.87\tiny{±0.11} & 0.86\tiny{±0.10} & 0.86\tiny{±0.11} & 0.82\tiny{±0.11} \\
possibly   & 0.50\tiny{±0.16} & 0.89\tiny{±0.10} & 0.50\tiny{±0.15} & 0.59\tiny{±0.18} & 0.84\tiny{±0.13} & 0.24\tiny{±0.17} & 0.46\tiny{±0.15} & 0.85\tiny{±0.17} & 0.86\tiny{±0.11} & 0.86\tiny{±0.09} & 0.85\tiny{±0.09} \\
\bottomrule
\end{tabular}}
\label{tab:mic_variance}
\end{table*}

\subsection{Full Results Excluding \texttt{<no\_hedge>}} \label{app:nohedge}
In \S\ref{subsec:mainresults}, we discuss the impact of excluding the special marker \texttt{<no\_hedge>} during metric computation and observe that \mrcx scores are negatively impacted by this change. We provide the full, unabbreviated results of this analysis setting, which are mostly consistent with our main conclusions, in Table \ref{tab:fullnohedge}.

\begin{table*}[t]
\centering\footnotesize
\setlength{\tabcolsep}{5pt}
\caption{Full analysis of \mic s when the special marker \texttt{<no\_hedge>} is excluded. $\Delta \mrc$ indicates the change in \mrcx after the exclusion, in comparison to the original \mrcx scores in Table \ref{tab:main}.}
\begin{tabular}{@{}lccccc >{\columncolor{blue!10}}ccc@{}}
\toprule
& \multicolumn{3}{c}{Marker Internal Confidence} & Density & \multicolumn{3}{c}{Rank}\\
\cmidrule(lr){2-4} \cmidrule(lr){5-5} \cmidrule(lr){6-9}
Model & \imaex $\downarrow$ & \cmaex $\downarrow$ & \mcvx $\downarrow$ & \dcvx & \mrcx$\uparrow$ & $\Delta \mrc$  & \macx & \mccx \\ \midrule
Gemini-2.5-Flash & 0.11 & 0.22 & 0.34 & 0.22 & \textbf{0.39} & $-0.47$ & 0.52 & 0.37 \\ 
Gemini-3-Flash & \textbf{0.08} & 0.14 & 0.18 & 0.14 & 0.03 & $-0.73$ & 0.33 & -0.70 \\
Gemini-3.1-Pro & \textbf{0.08} & \textbf{0.12} & \textbf{0.13} & 0.07 & 0.12 & $-0.20$ & 0.05 & -0.48 \\ \midrule
GPT-5-Nano & 0.15 & 0.26 & 0.32 & 0.25 & 0.36 & $-0.54$ & 0.23 & 0.35 \\ 
GPT-5-Mini & \textbf{0.11} & \textbf{0.19} & 0.27 & 0.29 & \textbf{0.41} & $-0.29$ & 0.43 & 0.05 \\ 
GPT-5 & \textbf{0.11} & 0.26 & \textbf{0.18} & 0.25 & 0.33 & $-0.50$ & 0.55 & -0.11 \\ \midrule
Qwen3-0.6B & 0.14 & 0.26 & 0.27 & 0.24 & 0.52 & $-0.17$ & 0.64 & 0.58 \\ 
Qwen3-1.7B & 0.12 & 0.21 & 0.26 & 0.24 & \textbf{0.78}& $+0.01$ & 0.60 & 0.17 \\ 
Qwen3-4B & \textbf{0.10} & 0.20 & 0.22 & 0.18 & 0.47& $-0.32$ & 0.53 & 0.20 \\ 
Qwen3-8B & \textbf{0.10} & 0.15 & \textbf{0.16} & 0.17 & 0.47 & $-0.38$ & 0.63 & 0.02 \\ 
Qwen3-32B & 0.11 & \textbf{0.17} & 0.21 & 0.20 & 0.27& $-0.61$ & 0.59 & -0.05 \\ \midrule
Llama3.1-8B-Instruct & 0.16 & 0.29 & 0.40 & 0.29 & 0.18& $-0.28$ & 0.61 & 0.33 \\ 
Llama3.3-70B-instruct & \textbf{0.14} & \textbf{0.21} & \textbf{0.21} & 0.19 & \textbf{0.80} & $+0.12$ & 0.38 & -0.46 \\ \bottomrule
\end{tabular}
\label{tab:fullnohedge}
\end{table*}

\subsection{Impact of $\mae$ Aggregation Granularity} \label{app:aggregation}
In \S\ref{app:metrics}, we discuss two alternate computational formulations of the \imaex and \cmaex based on sentence- and response-level aggregation of absolute errors between \mic s and observed intrinsic confidence, as opposed to the original marker-level aggregation. While these alternative formulations are susceptible to marker frequency bias, we still present comparative results to study the impact of aggregation granularity. For all main experimental settings, we compare the metric variants in Table \ref{tab:mae}, additionally reporting pooled standard deviation for each metric to capture estimate variability. It can be seen that choice of $\mae$ aggregation method has minimal impact on overall score, with consistent trends and comparable magnitudes observed across all variants. However, variance is higher for the sample- and response-level aggregation, and more generally, the observed variance suggests that \mic-confidence alignment and cross-domain generalizability of \mic s are inconsistent across tasks, sensitive to the specific datasets involved. This is in agreement with our main observations (\S\ref{subsec:mainresults}). 

\begin{table*}[h]
\centering\footnotesize
\setlength{\tabcolsep}{5pt}
\caption{Impact of $\mae$ aggregation level on \micx alignment results. Metric variants are described in \S\ref{app:metrics}.}
\begin{tabular}{@{}lcccccc@{}}
\toprule
& \multicolumn{3}{c}{\imaex Variants} & \multicolumn{3}{c}{\cmaex Variants}\\
\cmidrule(lr){2-4} \cmidrule(lr){5-7}
Model & \imaex $\downarrow$    &  $\imae_S \downarrow$   & $\imae_R \downarrow$   & \cmaex $\downarrow$   &  $\cmae_S \downarrow$   & $\cmae_R \downarrow$   \\ \midrule
Gemini-2.5-Flash & 0.11 $\pm$ 0.03 & 0.11 $\pm$ 0.08 & 0.11 $\pm$ 0.11 & 0.22 $\pm$ 0.06 & 0.19 $\pm$ 0.08 & 0.19 $\pm$ 0.09 \\ 
Gemini-3-Flash & 0.09 $\pm$ 0.04 & 0.09 $\pm$ 0.07 & 0.09 $\pm$ 0.06 & 0.13 $\pm$ 0.07 & 0.12 $\pm$ 0.09 & 0.12 $\pm$ 0.09 \\
Gemini-3.1-Pro & 0.08 $\pm$ 0.03 & 0.08 $\pm$ 0.05 & 0.08 $\pm$ 0.05 & 0.12 $\pm$ 0.06 & 0.12 $\pm$ 0.08 & 0.12 $\pm$ 0.08 \\ \midrule
GPT-5-Nano & 0.15 $\pm$ 0.02 & 0.17 $\pm$ 0.09 & 0.13 $\pm$ 0.12 & 0.26 $\pm$ 0.05 & 0.24 $\pm$ 0.09 & 0.24 $\pm$ 0.08 \\ 
GPT-5-Mini & 0.11 $\pm$ 0.06 & 0.13 $\pm$ 0.10 & 0.13 $\pm$ 0.11 & 0.19 $\pm$ 0.05 & 0.17 $\pm$ 0.09 & 0.17 $\pm$ 0.07 \\
GPT-5 & 0.11 $\pm$ 0.04 & 0.14 $\pm$ 0.09 & 0.14 $\pm$ 0.10 & 0.26 $\pm$ 0.01 & 0.28 $\pm$ 0.12 & 0.27 $\pm$ 0.13 \\ \midrule
Qwen3-0.6B & 0.14 $\pm$ 0.04 & 0.18 $\pm$ 0.10 & 0.13 $\pm$ 0.13 & 0.26 $\pm$ 0.06 & 0.26 $\pm$ 0.09 & 0.26 $\pm$ 0.04 \\ 
Qwen3-1.7B & 0.12 $\pm$ 0.03 & 0.14 $\pm$ 0.09 & 0.12 $\pm$ 0.11 & 0.21 $\pm$ 0.07 & 0.19 $\pm$ 0.07 & 0.19 $\pm$ 0.07 \\ 
Qwen3-4B & 0.10 $\pm$ 0.03 & 0.12 $\pm$ 0.08 & 0.11 $\pm$ 0.10 & 0.20 $\pm$ 0.07 & 0.17 $\pm$ 0.09 & 0.17 $\pm$ 0.07 \\ 
Qwen3-8B & 0.10 $\pm$ 0.03 & 0.11 $\pm$ 0.09 & 0.11 $\pm$ 0.12 & 0.15 $\pm$ 0.03 & 0.16 $\pm$ 0.10 & 0.15 $\pm$ 0.09 \\ 
Qwen3-32B & 0.11 $\pm$ 0.02 & 0.11 $\pm$ 0.07 & 0.10 $\pm$ 0.10 & 0.13 $\pm$ 0.04 & 0.15 $\pm$ 0.08 & 0.15 $\pm$ 0.06 \\ \midrule
Llama3.1-8B-Instruct & 0.16 $\pm$ 0.03 & 0.19 $\pm$ 0.10 & 0.14 $\pm$ 0.13 & 0.29 $\pm$ 0.06 & 0.28 $\pm$ 0.10 & 0.28 $\pm$ 0.09 \\ 
Llama3.3-70B-instruct & 0.14 $\pm$ 0.03 & 0.13 $\pm$ 0.09 & 0.10 $\pm$ 0.09 & 0.21 $\pm$ 0.06 & 0.15 $\pm$ 0.09 & 0.14 $\pm$ 0.04 \\ \bottomrule
\end{tabular}
\label{tab:mae}
\vspace{-4mm}
\end{table*}

\subsection{Impact of Correlation Type on \macx and \mcc} \label{app:corr}
We present additional results of using Spearman correlations instead of Pearson correlations to compute \macx and \mccx for all main experimental settings in Table \ref{tab:corr}. It can be seen that results are fairly consistent, suggesting that the linear proportionality and ranking order of \mic s with accuracy or faithful calibration level for most models is robust to the choice of correlation measure and holds under both linear and rank-based assumptions.

\begin{table*}[t]
\vspace{5mm}
\centering\footnotesize
\setlength{\tabcolsep}{10pt}
\caption{Impact of correlation coefficient type on \macx and \mccx results. Metric variants are described in \S\ref{app:metrics}.}
\begin{tabular}{@{}lcccc@{}}
\toprule
Model & $\mac$ & $\mac_{\text{Spear}}$ & $\mcc$ & $\mcc_{\text{Spear}}$ \\ \midrule
Gemini-2.5-Flash & 0.47 & 0.47 & 0.35 & 0.29 \\ 
Gemini-3-Flash & 0.26 & 0.19 & -0.66 & -0.68 \\
Gemini-3.1-Pro & 0.07 & 0.29 & -0.49 & -0.58 \\ \midrule
GPT-5-Nano & 0.62 & 0.54 & -0.40 & -0.42 \\ 
GPT-5-Mini & 0.60 & 0.56 & -0.02 & 0.00 \\ 
GPT-5 & 0.63 & 0.68 & -0.21 & -0.14 \\ \midrule
Qwen3-0.6B & 0.70 & 0.70 & 0.68 & 0.71 \\ 
Qwen3-1.7B & 0.51 & 0.47 & 0.15 & 0.24 \\ 
Qwen3-4B & 0.61 & 0.61 & 0.23 & 0.11 \\ 
Qwen3-8B & 0.65 & 0.65 & -0.24 & -0.33 \\ 
Qwen3-32B & 0.66 & 0.59 & -0.11 & -0.04 \\ \midrule
Llama3.1-8B-Instruct & 0.69 & 0.61 & 0.30 & 0.30 \\ 
Llama3.3-70B-Instruct & 0.34 & 0.34 & -0.45 & -0.57 \\ \bottomrule
\end{tabular}
\label{tab:corr}
\vspace{-4mm}
\end{table*}

\end{document}